\documentclass{article} % For LaTeX2e
\usepackage{iclr2027_conference,times}

\usepackage{amsmath,amsfonts,bm}

\def\eqref#1{equation~\ref{#1}}
\def\1{\bm{1}}

\DeclareMathAlphabet{\mathsfit}{\encodingdefault}{\sfdefault}{m}{sl}
\SetMathAlphabet{\mathsfit}{bold}{\encodingdefault}{\sfdefault}{bx}{n}

\usepackage[utf8]{inputenc} % allow utf-8 input
\usepackage[T1]{fontenc}    % use 8-bit T1 fonts
\usepackage{hyperref}       % hyperlinks
\usepackage{url}            % simple URL typesetting
\usepackage{booktabs}       % professional-quality tables
\usepackage{amsfonts}       % blackboard math symbols
\usepackage{nicefrac}       % compact symbols for 1/2, etc.
\usepackage{microtype}      % microtypography
\usepackage{xcolor}         % colors
\usepackage[table]{xcolor}
\usepackage{algorithm}
\usepackage{algpseudocode}
\usepackage{amsmath}
\usepackage{graphicx} 
\usepackage{multirow}
\usepackage{hyperref}
\usepackage{array,booktabs}
\newcolumntype{L}[1]{>{\raggedright\arraybackslash}m{#1}}
\newcolumntype{C}[1]{>{\centering\arraybackslash}m{#1}}

\usepackage{tabularx}
\usepackage{array}
\usepackage{textcomp}
\usepackage{amssymb} 
\usepackage{makecell}
\usepackage{pifont}
\newcolumntype{Y}{>{\centering\arraybackslash}X}
\newcommand{\cmark}{\textcolor{green!55!black}{\ding{51}}}
\newcommand{\xmark}{\textcolor{red!75!black}{\ding{55}}}
\newcommand{\citenumonly}[1]{[\citenum{#1}]}
\definecolor{basegray}{gray}{0.4}
\usepackage{float}

\title{Bottlenecked Semantic Compression by
Representation Reconstruction}

\author{Xiaoya Tang \\
Scientific Computing and Imaging Institute \\
University of Utah \\
\texttt{xiaoya.tang@utah.edu}
\And
Shilong Liu\textsuperscript{\textdagger} \\
Princeton AI Lab \\
Princeton University \\
\texttt{shilong.liu@princeton.edu}
\And
Daniel Smolyak\textsuperscript{\textdagger} \\
National Institutes of Health \\
\texttt{daniel.smolyak@nih.gov}
\AND
Xiwen Li \\
Scientific Computing and Imaging Institute \\
University of Utah \\
\texttt{xiwen.li@utah.edu}
\And
Quynh Nguyen \\
National Institutes of Health \\
\texttt{quynh.nguyen3@nih.gov}
\And
Tolga Tasdizen\textsuperscript{*} \\
Scientific Computing and Imaging Institute \\
University of Utah \\
\texttt{tolga.tasdizen@utah.edu}
\\[1ex]
\textsuperscript{\textdagger}Equal Contribution.
\qquad
\textsuperscript{*}Corresponding authors.
}

\iclrfinalcopy % Uncomment for camera-ready version, but NOT for submission.

\begin{document}
\maketitle
\lhead{Under review as a conference paper at ICLR 2027}

\begin{abstract}
Vision Transformers incur high computational costs when processing long visual token sequences. Existing reduction methods typically rank tokens independently and discard those deemed unimportant or redundant. We propose a Reconstruction-guided framework for visual token condensation(ReConV), which that summarizes the complete visual sequence with a compact Summary Bank (SMB) and retains tokens whose information is not adequately captured by the SMB, as indicated by their reconstruction residuals. At each compression layer, a lightweight position-conditioned decoder reconstructs the current tokens, leaving shared information compressed in the SMB while explicitly preserving poorly reconstructed details. Experiments on image classification and multimodal understanding demonstrate favorable performance--efficiency trade-offs: ReConV surpasses the uncompressed baseline with \(48\%\) fewer GMACs on ImageNet and enables reusable image-level representations for efficient multi-turn inference. Code is available at \url{TokenDrop}.
\end{abstract}

\section{Introduction}\label{sec:intro}
Vision Transformers (ViTs) have become a dominant architecture in modern computer vision, but their self-attention cost grows quadratically with the number of input tokens. This quickly becomes prohibitive for high-resolution images, multiple images, and videos. The problem is even more severe in multimodal large language models (MLLMs), where visual tokens are concatenated with text tokens and jointly processed by an LLM, leading to memory, latency, and throughput burdens.

The redundancy of visual representations has motivated token reduction methods that prune or merge tokens under a constrained computational budget. In conventional visual recognition, token reduction is motivated by the observation that image understanding progressively condenses local patch information into higher-level semantic representations. Accordingly, existing methods estimate token importance~\cite{rao2021dynamicvit,yin2022avit}, merge tokens with similar representations~\cite{bolya2023tome}, or formulate reduction as a clustering problem~\cite{chang2023making}. Most of these methods, however, were developed for single-image classification and are often guided by object-centric discriminative cues. Such assumptions are less suitable for MLLMs, where visual tokens must additionally preserve spatial grounding and information that may become relevant only when combined with a textual instruction.

\begin{figure*}[!t]
    \centering
    \includegraphics[width=0.85\textwidth]{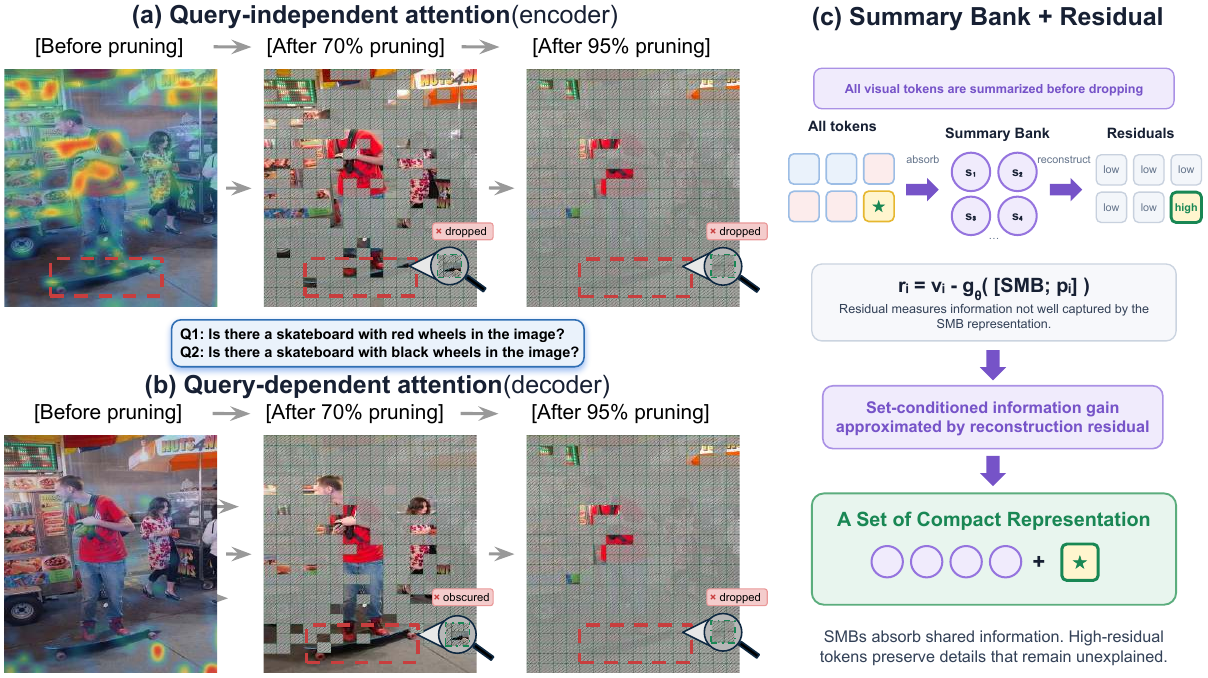}
    \caption{
    Illustration of visual token pruning.
    Prompt-independent saliency emphasizes irrelevant regions, while prompt-conditioned pruning can still discard task-relevant details, especially at high pruning ratios.
    ReConV instead preserves information through SMB summaries and residual tokens.
    }
    \label{fig:into_attention_illustration}
\end{figure*}

Consequently, visual token reduction in MLLMs has increasingly relied on text-to-image attention from either the vision encoder or the LLM decoder to rank visual tokens~\cite{shang2025llavaprumerge,chen2024image,xing2024pyramiddrop,chen2024fastv,chen2026accelerating}. However, attention primarily reflects local information routing for a particular layer, head, and query, with no guarantee that its magnitude reliably captures token utility or the causal effect of removing a token. It may also be affected by model-induced saliency priors and systematic preferences for particular tokens, and vary across layers and heads. Moreover, accessing attention matrices can interfere with efficient kernels~\cite{dao2024flashattention}, while decoder-side pruning cannot reduce computation in preceding LLM layers. We summarizes representative visual token reduction paradigms in Table~\ref{tab:intro-checkbox}. To eliminate dependence on attention, we introduce a learnable Summary Bank (SMB) that semantically aggregates the complete visual sequence, and a lightweight position-conditioned decoder that reconstructs the original tokens from this compact representation. As illustrated in Figure~\ref{fig:into_attention_illustration}, this reconstruction-based criterion requires neither attention distributions nor textual prompts and enables visual token reduction before the LLM.

\begin{table}
\centering
\caption{
Comparison of representative visual token reduction paradigms.
A \cmark\ denotes explicit design support, whereas an \xmark\ denotes no explicit guarantee.
\emph{No attention-layer/head dependence} excludes reliance on specific attention layers or heads; \emph{prompt independence} excludes textual conditioning; \emph{reduction before the LLM} means compression precedes LLM processing; \emph{preserving uncaptured information} retains information missing from the compressed representation; and \emph{joint token information coverage} evaluates tokens relative to the selected set rather than independently.
}
\label{tab:intro-checkbox}
\setlength{\tabcolsep}{5pt}
\renewcommand{\arraystretch}{1.00}
\footnotesize
\resizebox{\textwidth}{!}{
\begin{tabular}{lccccc}
\toprule
\textbf{Method}
&
\makecell{\textbf{No attention-layer}\\
          \textbf{/ head dependence}}
&
\makecell{\textbf{Prompt}\\
          \textbf{independence}}
&
\makecell{\textbf{Reduction before}\\
          \textbf{the LLM}}
&
\makecell{\textbf{Preserves uncaptured}\\
          \textbf{information}}
&
\makecell{\textbf{Joint token redundancy}\\
          \textbf{/ coverage}}
\\
\midrule
Decoder-attention ranking
& \xmark & \xmark & \xmark & \xmark & \xmark \\
Encoder-attention ranking
& \xmark & \cmark & \cmark & \xmark & \xmark \\
Other learned scoring / gating
& \cmark & \cmark & \cmark & \xmark & \xmark \\
Similarity-based clustering / merging
& \cmark & \cmark & \cmark & \xmark & \cmark \\
\midrule
\makecell[l]{\textbf{Summary Bank + residual}\textbf{(Ours)}}
& \cmark & \cmark & \cmark & \cmark & \cmark \\
\bottomrule
\end{tabular}
}
\end{table}

A second limitation is that existing scoring methods evaluate tokens independently, potentially retaining redundant high-ranking tokens while discarding subtle evidence. Although merging~\cite{shang2025llavaprumerge} accounts for redundancy, it may obscure details absent from the compressed representation. ReConV instead encodes the complete visual sequence into the SMBs and uses reconstruction residuals to identify uncaptured information: well-reconstructed tokens are removed, while high-residual tokens are retained as complementary evidence. Selection therefore depends on what the SMB has jointly captured, bridging pruning and information aggregation. Our method retains \(100.1\%\) of the baseline classification performance with \(48\%\) fewer GMACs and \(89\%\) of multi-turn performance using only \(5\%\) of the visual tokens. Patch-to-SMB assignments also reveal emergent semantically coherent aggregation (Figure~\ref{fig:smb_vis_cls}).

Our main contributions are as follows:
\begin{enumerate}
\setlength{\itemsep}{0pt}
\setlength{\parsep}{0pt}
\setlength{\parskip}{0pt}
\setlength{\topsep}{2pt}
\setlength{\partopsep}{0pt}
\item We propose ReConV, a reconstruction-guided framework that compresses visual tokens through a compact Summary Bank (SMB) and lightweight position-conditioned reconstruction.
% \item We use reconstruction residuals to retain information not jointly captured by the SMB, replacing independent token scoring with set-conditioned selection. Patch-to-SMB assignments further reveal emergent semantic grouping. 
\item We propose a set of learnable Summary Bank (SMB) tokens that aggregate the complete visual sequence through asymmetric cross-attention. Reconstruction residuals measure information not captured by this joint summary and guide token retention. Patch-to-SMB assignments further reveal emergent semantic grouping.
% \item We propose a set of learnable Summary Bank (SMB) tokens that aggregate the complete visual sequence through asymmetric cross-attention. Reconstruction residuals identify information not captured by this joint summary, removing low-residual tokens while retaining high-residual ones rather than relying on independent importance scores. Patch-to-SMB assignments further reveal emergent semantically coherent grouping.
\item ReConV generalizes across classification and multimodal dialogue, matching the uncompressed classification baseline with \(48\%\) fewer GMACs and enabling efficient multi-turn inference through reusable image-level representations.
\end{enumerate}

\section{Related Work}\label{sec:Related Work}

% \subsection{Token Reduction in Classification}\label{sec:related-token-reduc-cls}
Token reduction in ViTs is motivated by the observation that accurate
recognition often requires only a subset of the visual tokens. Early methods
such as DynamicViT~\cite{rao2021dynamicvit}, EViT~\cite{liang2022evit}, Evo-ViT~\cite{xu2022evovit}, A-ViT~\cite{yin2022avit}, and ATS~\cite{fayyaz2022ats} estimate token importance using attention signals or learned scoring functions and remove low-scoring tokens during inference. To preserve information that would
otherwise be discarded, subsequent methods such as PPT~\cite{wu2023ppt}, ToMe~\cite{bolya2023tome},
Token Pooling~\cite{marin2023tokenpooling}, LookupViT~\cite{koner2024lookupvit}, and STViT~\cite{chang2023making} compress tokens through merging, pooling, or clustered latent representations. Our method instead uses reconstruction-guided Summary Bank (SMB) tokens as a latent bottleneck, avoiding direct importance scoring and irreversible merging. 
% While related to Perceiver~\cite{jaegle2021perceiver} and Perceiver IO~\cite{jaegle2021perceiverio}, which compress inputs via asymmetric cross-attention, our framework uses reconstruction residuals to preserve information not captured by the bottleneck.

% \subsection{Visual Token Reduction in MLLMs}\label{sec:related-token-reduc-mllm}
Visual token reduction in MLLMs is often text-guided. SparseVLM~\cite{zhang2024sparsevlm} uses decoder cross-attention and recycles pruned tokens through density-peak aggregation, while FastV~\cite{chen2024image} and PDrop~\cite{xing2024pyramiddrop} prune according to LLM or instruction-to-image attention. PruMerge~\cite{shang2025llava} and VisionZip~\cite{yang2025visionzip} employ merging, whereas AdaptVision~\cite{lin2025adaptvision} selectively acquires high-resolution crops. In contrast, our vision-only, reconstruction-guided framework requires neither prompts, attention heuristics, nor region-grounding supervision, enabling reusable image-level compression across ViTs and MLLMs.

Despite their differences, these methods rely on two approximations: \emph{attention magnitude is a reliable proxy for token utility}, and \emph{independent top-\(K\) scoring approximates subset optimization}, although a token's utility depends on what has already been preserved. Consequently, redundant high-scoring tokens may be retained while complementary evidence is discarded (Fig.~\ref{fig:into_attention_illustration}). We address both limitations with an SMB that summarizes the complete sequence and reconstruction residuals that measure each token's uncaptured information relative to this summary. Sec.~\ref{sec:method} details the method.

\section{Methodology}\label{sec:method}
\begin{figure*}[t]
    \centering
    \includegraphics[width=\textwidth]{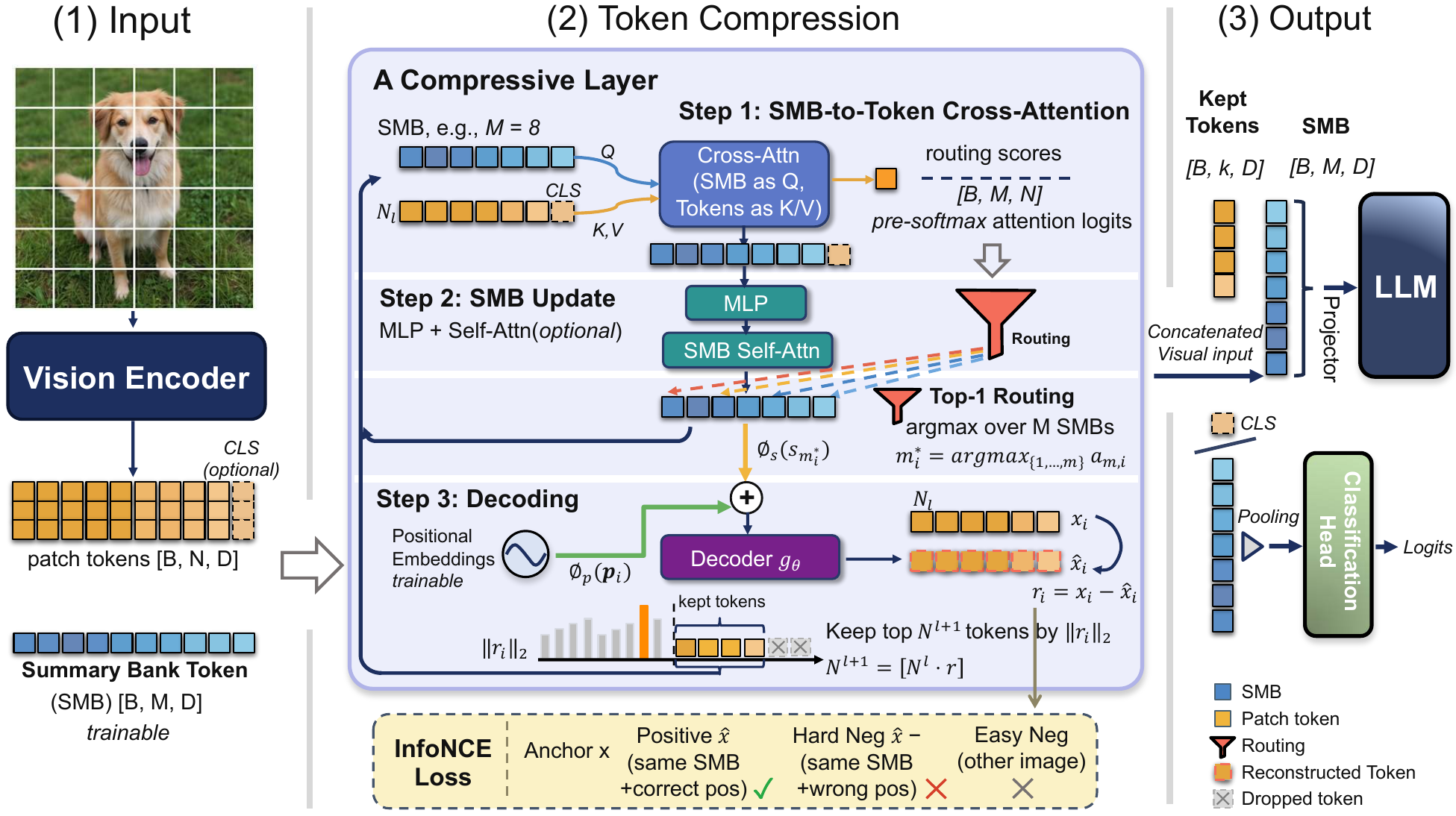}
    \caption{Overview of ReConV, the proposed reconstruction-guided token pruning framework. Given $N$ token embeddings from a vision encoder or transformer layer, each compressive layer uses a small set of learned summary bank (SMB) tokens to summarize the input via cross-attention, reconstructs the original tokens using the SMB representation and positional information, and retains the top-$k$ tokens with the largest reconstruction residuals (approximately $N\cdot r$, where $r$ is the keep ratio). Tokens that are well reconstructed are dropped as redundant, while the retained tokens are passed to the next layer. For classification, the model uses the CLS token when available, or pooled SMB features otherwise; for other tasks, it uses the retained visual tokens with SMBs.}
    \label{fig:methodology}
\end{figure*}

% \subsection{Reconstruction-Guided Compression}\label{sec:method-reconguidedcomp}
Given visual token embeddings $\mathbf{V}^{(\ell)}
\in \mathbb{R}^{B\times N_\ell\times D}$ from a standard vision transformer layer, where $N$ denotes the original token length before compression, our compressive layers learn a small set of \emph{summary bank} (SMB) tokens that summarize the original token set, and discard only those tokens that are already well explained by this learned summary while forwarding the retained tokens to the next layer. Each compressive layer takes the current token set, the current SMB, and positional embeddings as input, produces an updated SMB and a reduced token set, reconstructs tokens with a two-layer MLP decoder, and retains only the top-$k$ tokens according to a residual-based criterion, where $k$ is determined by a prescribed keep ratio. The retained token embeddings are then forwarded to the next layer. Compressive layers can be inserted progressively between transformer encoder blocks, trained from scratch with normal encoder layers or applied as a separate module after a pretrained and frozen vision encoder, as in our MLLM setting. We next describe a single compressive layer in detail, shown in Figure~\ref{fig:methodology}. The pseudo-code for a compressive layer is provided in Appendix~\ref{appendix:1}.

\subsection{Summary Bank}\label{sec:method-smb}
Let $\mathbf{V}^{(\ell-1)}
\in\mathbb{R}^{B\times N_{\ell-1}\times D}$
denote the visual token sequence entering the $\ell$-th compressive layer,
and let $\mathbf{S}^{(\ell-1)}
\in\mathbb{R}^{B\times M\times D}$
denote the Summary Bank (SMB) propagated from the preceding compressive layer,
where $N_0=N$, $1\leq N_{\ell-1}\leq N$ and $M\ll N$.
The SMB is initialized from a learnable parameter
$\mathbf{S}_0\in\mathbb{R}^{1\times M\times D}$, broadcast along the batch
dimension. Each compressive layer updates the SMB through asymmetric
cross-attention, using the SMB tokens as queries and the visual tokens as
keys and values:
\begin{equation}
\widetilde{\mathbf{S}}^{(\ell)}
=
\mathbf{S}^{(\ell-1)}
+
\mathrm{MHA}\!\left(
    \mathrm{LN}(\mathbf{S}^{(\ell-1)}),
    \mathrm{LN}(\mathbf{V}^{(\ell-1)}),
    \mathrm{LN}(\mathbf{V}^{(\ell-1)})
\right).
\label{eq:method-SMB}
\end{equation}

After cross-attention, we further refine $\widetilde{\mathbf{S}}^{(\ell)}$ with an MLP, optionally followed by an SMB-side multi-head self-attention block. When a class token is present, it participates only in these refinement operations and is excluded from SMB-to-visual cross-attention and token reconstruction. We denote the resulting summary bank by $\mathbf{S}^{(\ell)}$. In the following subsections, we omit the layer index and use $\mathbf{S}$ and $\mathbf{s}_m$ to denote the updated summary bank and its $m$-th token, respectively.

\subsection{Token Routing and Reconstruction}\label{sec:method-decoding}
The pre-softmax attention logits from SMB-to-token cross-attention, averaged across heads, are reused as routing scores. For each visual token $\mathbf{v}_i$, we select the summary slot with the highest score:
\begin{equation}
m_i^\star
=
\arg\max_{m\in\{1,\ldots,M\}} a_{m,i},
\label{eq:method-argmaxSMB}
\end{equation}
where $a_{m,i}$ denotes the attention logit between the $m$-th SMB token $\mathbf{s}_m$ and visual token $\mathbf{v}_i$. This top-1 routing assigns each visual token to a single SMB token, requiring only one decoder evaluation per token. To preserve token-level spatial identity, we maintain a learnable positional table
$\mathbf{Pos}\in\mathbb{R}^{1\times N\times D}$. The original position indices of retained tokens are propagated across compressive layers, allowing the corresponding positional embedding $\mathbf{p}_i$ to be retrieved at each layer. Given the routed SMB token $\mathbf{s}_{m_i^\star}$ and positional embedding $\mathbf{p}_i$, we project both into the decoder space, concatenate them, and reconstruct the input token using a decoder $g_\theta$:
\begin{equation}
\hat{\mathbf{v}}_i
=
g_\theta\!\left(
    [\phi_s(\mathbf{s}_{m_i^\star});
     \phi_p(\mathbf{p}_i)]
\right).
\label{eq:method-decoding}
\end{equation}

The reconstructed token $\hat{\mathbf{v}}_i$ is used to compute the reconstruction residual for token selection and to define the training objective described in Section~\ref{sec:method-recon_loss}. Although no semantic supervision is imposed on the routing assignments, visualizing them reveals the spontaneous emergence of semantically coherent patch groups, as shown in Figure~\ref{fig:smb_vis_cls}. Additional examples are provided in Appendices~\ref{sec:apdx-mllm-smb-vis} and~\ref{sec:apdx-cls-smb-vis}.

\subsection{Reconstruction Objective and Token Selection}\label{sec:method-recon_loss}
We optimize the compression module with an InfoNCE reconstruction objective, while token selection is guided by the orthogonal residual derived from each reconstructed token. Given the reconstructed token $\hat{\mathbf{v}}_i$, we compute its normalized direction $\mathbf{u}_i$ and define the orthogonal residual for token selection as:
\begin{equation}
\mathbf{u}_i
=
\frac{\hat{\mathbf{v}}_i}
     {\|\hat{\mathbf{v}}_i\|_2},
\qquad
\mathbf{r}_i
=
\mathbf{v}_i
-
(\mathbf{v}_i^\top\mathbf{u}_i)\mathbf{u}_i.
\label{eq:method-infonce1}
\end{equation}

The token-selection score is given by $|\mathbf{r}_i|_2$. The top-$k$ tokens with the largest residual norms are retained, while the remaining tokens are dropped. Intuitively, this preserves tokens that remain poorly explained after removing the component aligned with their reconstructions. The reconstruction is optimized with a contrastive objective. Specifically, we normalize the original and reconstructed tokens and define the positive and negative contrastive logits as

\begin{equation}
\begin{gathered}
\bar{\mathbf{v}}_i
=
\frac{\mathbf{v}_i}{\|\mathbf{v}_i\|_2},
\quad
\bar{\hat{\mathbf{v}}}_j
=
\frac{\hat{\mathbf{v}}_j}{\|\hat{\mathbf{v}}_j\|_2},\qquad
\ell_i^{+}
=
\frac{\bar{\mathbf{v}}_i^\top\bar{\hat{\mathbf{v}}}_i}{\tau},
\quad
\ell_{i,j}^{-}
=
\frac{\bar{\mathbf{v}}_i^\top\bar{\hat{\mathbf{v}}}_j}{\tau},
\quad j\in\mathcal{N}_i,
\end{gathered}
\label{eq:method-infonce2}
\end{equation}
% we set
% \[
% \Omega=\{1,\dots,N\}\setminus\mathcal{I}_{\mathrm{keep}}.
% \]
where $\tau$ is the contrastive temperature and $\mathcal{N}_i$ denotes the negative set associated with anchor $i$. Anchors are sampled from a candidate pool $\Omega$, defined as either the dropped-token set or the full token set. We sample an anchor set $\mathcal{T}\subseteq\Omega$ uniformly without replacement, with
\begin{equation}
|\mathcal{T}|
=
\begin{cases}
|\Omega|, & T_0= 0,\\
\min(T_0,|\Omega|), & T_0>0,
\end{cases}
\label{eq:method-infonce3}
\end{equation}
where $T_0\geq 0$ is a pre-specified anchor budget, with $T_0=0$ indicating that all candidates in $\Omega$ are used. For each anchor $i$, hard negatives are reconstructed tokens at different positions within the same image that are routed to the same SMB slot, whereas easy negatives are reconstructed tokens from other images in the batch. 
The resulting reconstruction loss is
\begin{equation}
\mathcal{L}_{\mathrm{rec}}^{\mathrm{InfoNCE}}
=
-\frac{1}{|\mathcal{T}|}
\sum_{i\in\mathcal{T}}
\log
\frac{\exp(\ell_i^{+})}
{\exp(\ell_i^{+})
+\sum_{j\in\mathcal{N}_i}\exp(\ell_{i,j}^{-})}.
\label{eq:method-infonce4}
\end{equation}
Thus, the same reconstruction serves both token selection and module optimization: the orthogonal residual determines which tokens are retained, while the InfoNCE objective trains the compression module to recover token-specific representations. At compressive layer $\ell$, we retain
$k_\ell=\mathrm{round}(N_{\ell-1}\rho_\ell)$ tokens with the largest residual norms, where $\rho_\ell$ is the prescribed keep ratio. The retained representations forwarded to the next layer can be either the original token embeddings $\mathbf{v}_i$ or their residuals $\mathbf{r}_i$. Forwarding the residuals explicitly propagates complementary information not captured by the SMB reconstruction. Across $L$ compressive layers, the number of retained non-SMB tokens is approximately
\begin{equation}
N_{\mathrm{ret}}
\approx
N\prod_{\ell=1}^{L}\rho_\ell.
\label{eq:method-infonce5}
\end{equation}

When the SMB is included in the final compressed representation, its total length is approximately $M+N_{\mathrm{ret}}$.
Finally, the model is trained by combining the task and reconstruction objectives:
\begin{equation}
\mathcal{L}
=
\mathcal{L}_{\mathrm{task}}
+
\lambda\,
\mathcal{L}_{\mathrm{rec}}^{\mathrm{InfoNCE}},
\label{eq:method-infonce6}
\end{equation}
where $\lambda$ controls the contribution of the reconstruction objective.

\begin{figure*}[t]
\centering
\begingroup
\setlength{\tabcolsep}{4pt}
\resizebox{\textwidth}{!}{%
\begin{tabular}{cccccc}
\includegraphics[width=0.15\textwidth]{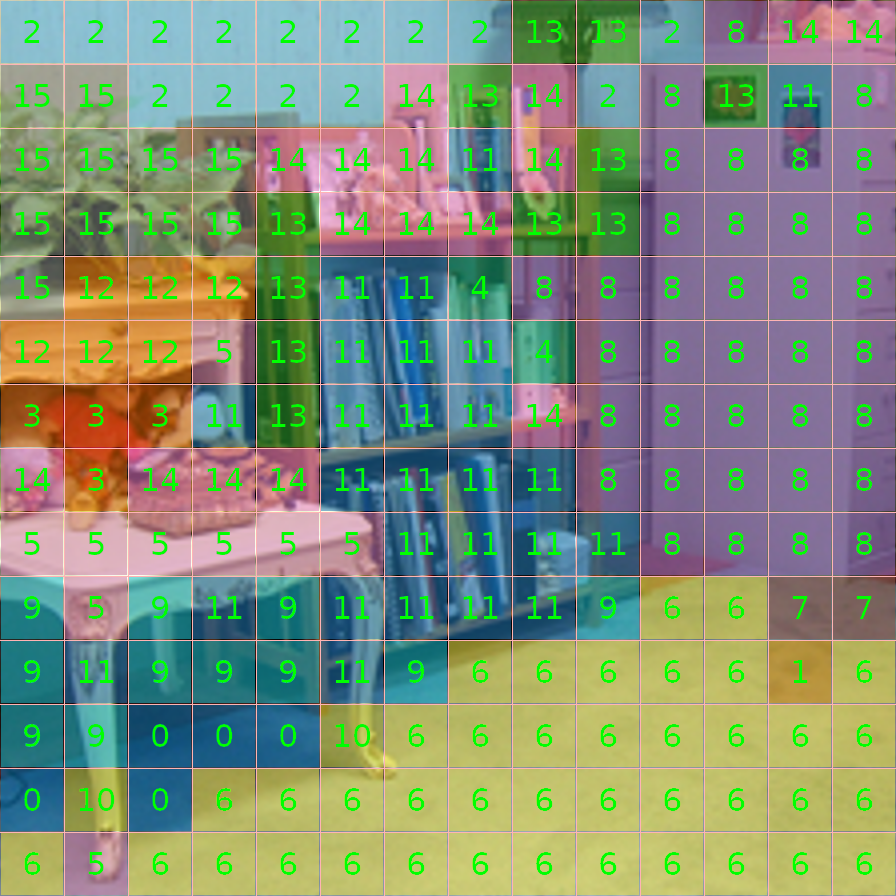} &
\includegraphics[width=0.15\textwidth]{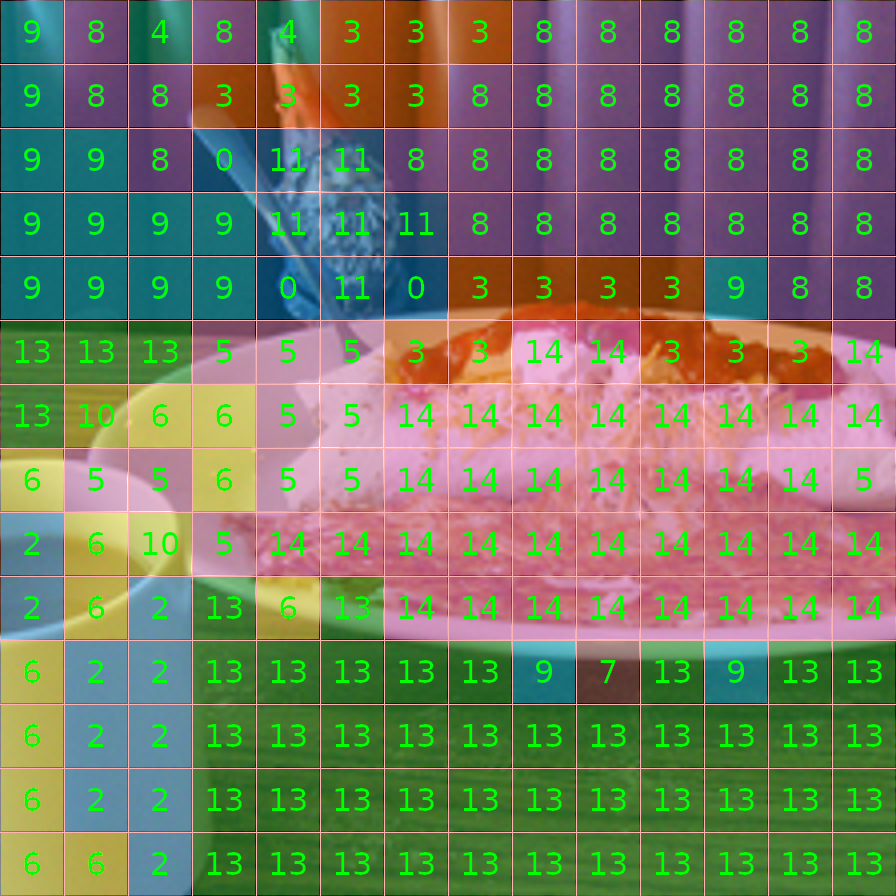} &
\includegraphics[width=0.15\textwidth]{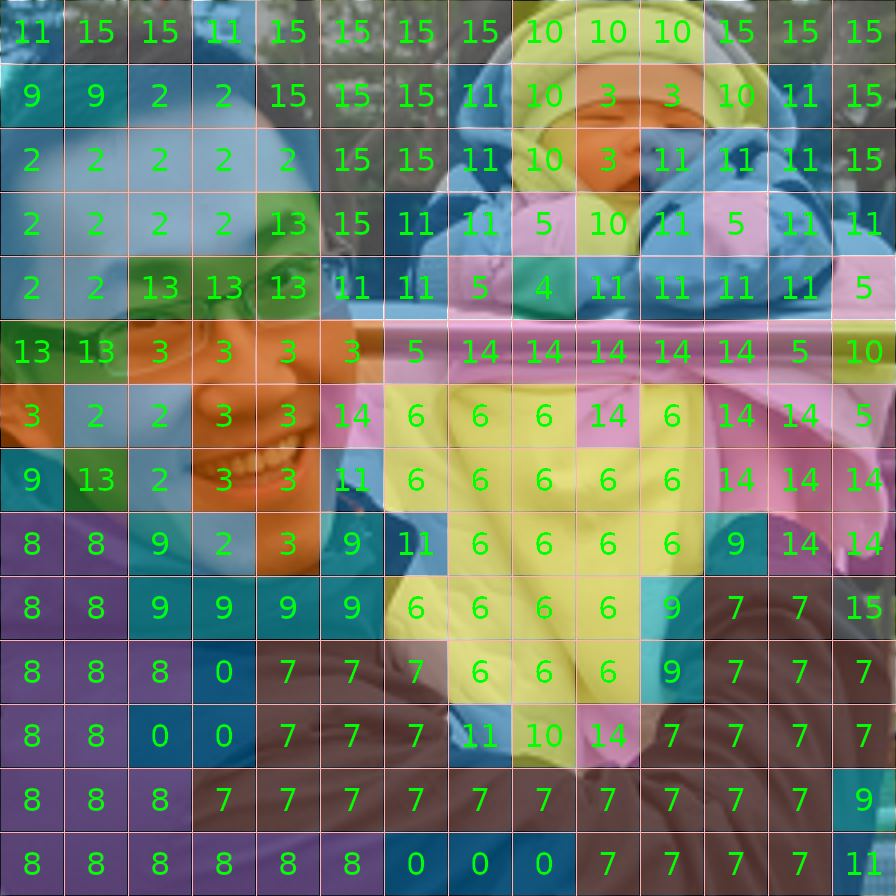} &
\includegraphics[width=0.15\textwidth]{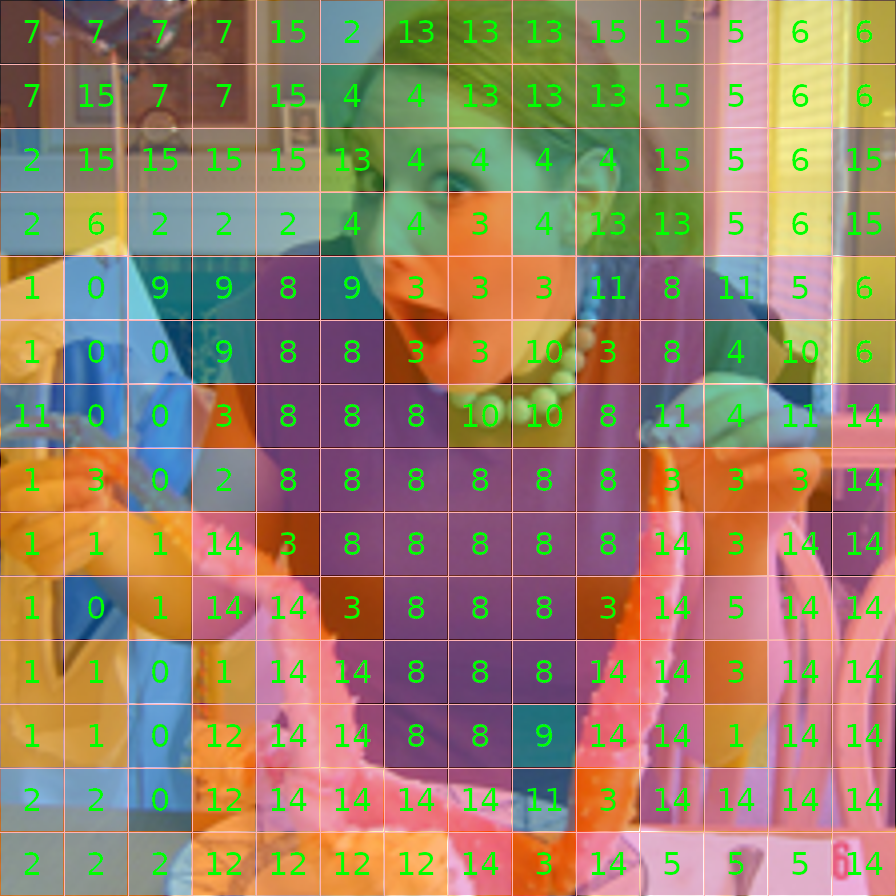} &
\includegraphics[width=0.15\textwidth]{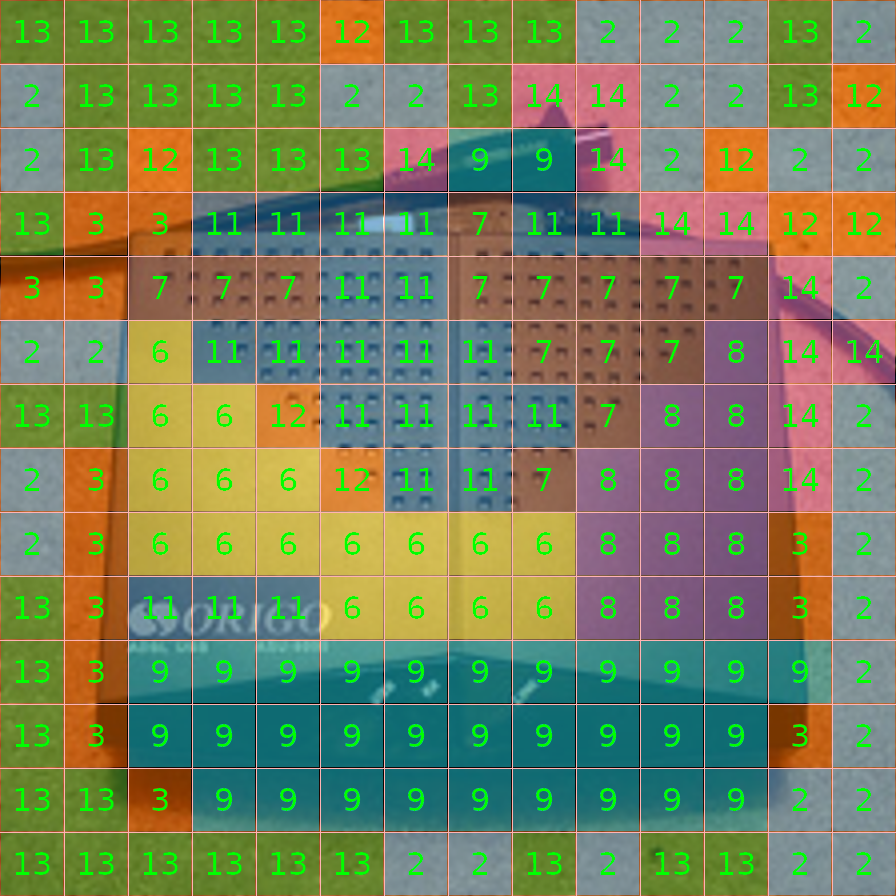} &
\includegraphics[width=0.15\textwidth]{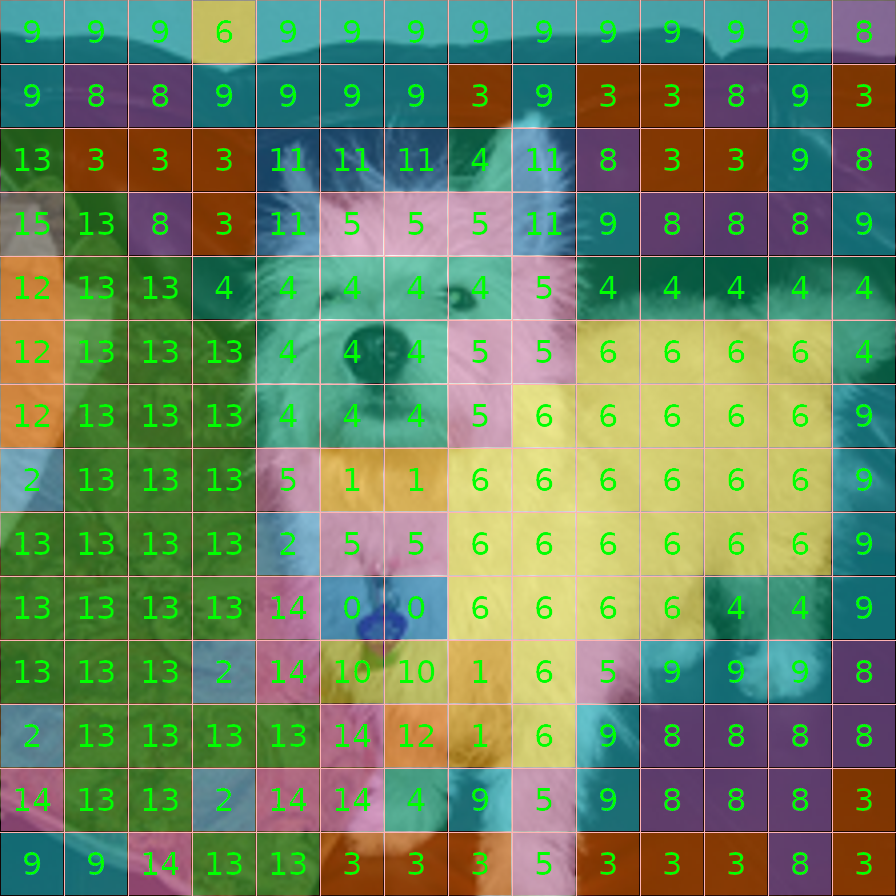}\\
\includegraphics[width=0.15\textwidth]{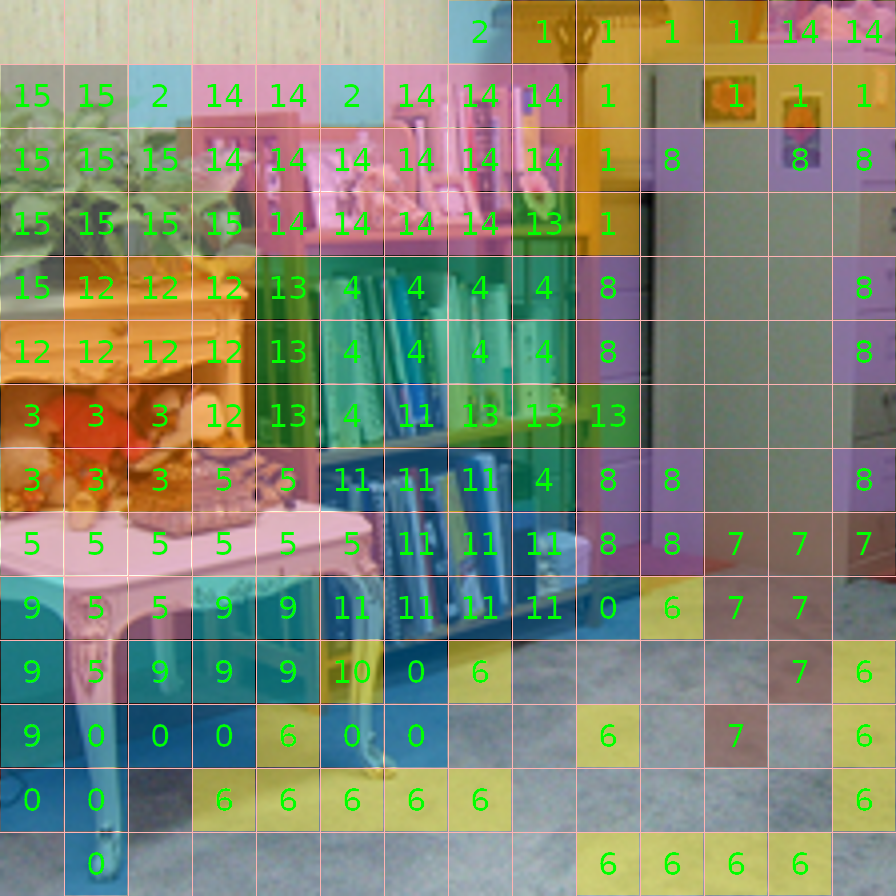} &
\includegraphics[width=0.15\textwidth]{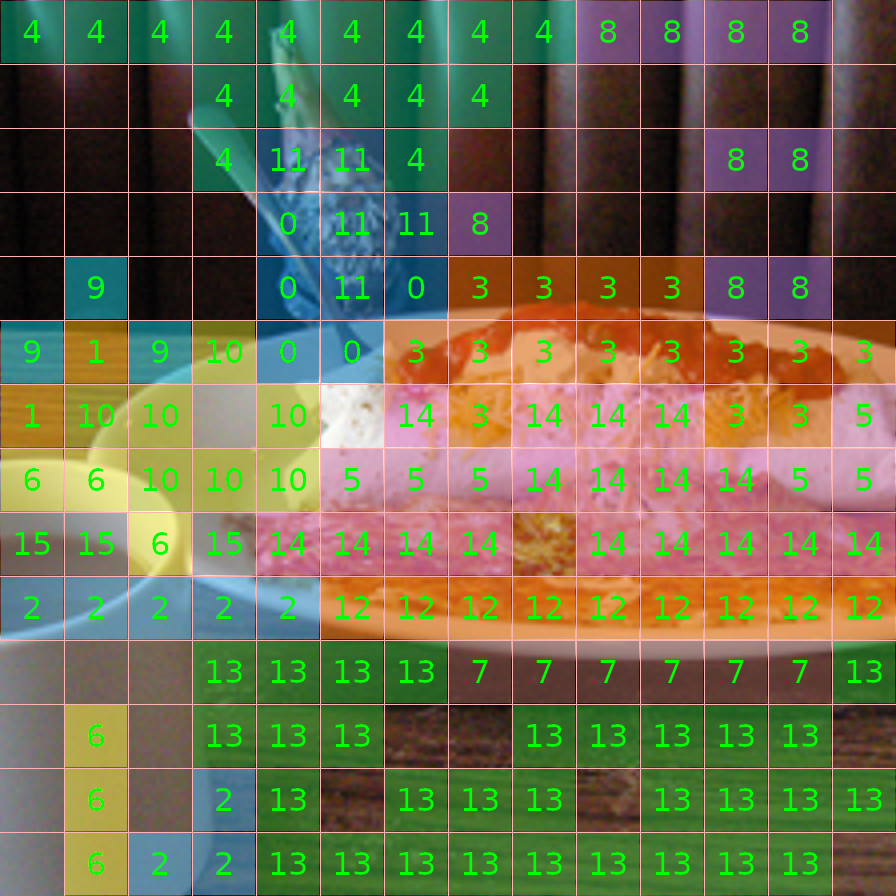} &
\includegraphics[width=0.15\textwidth]{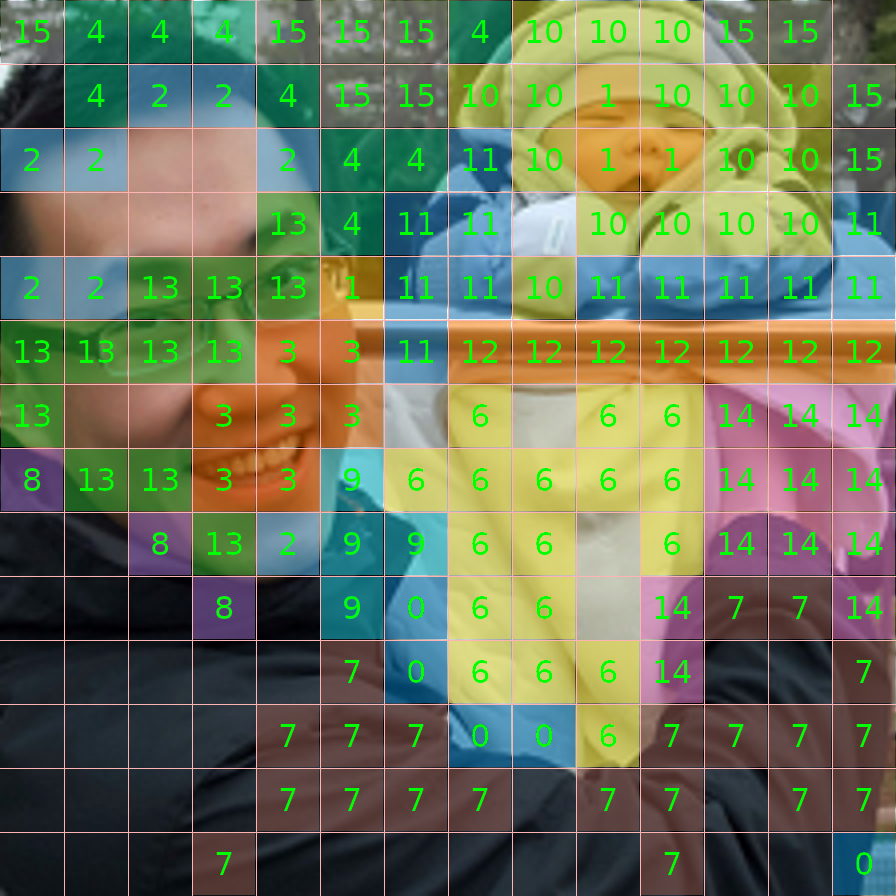} &
\includegraphics[width=0.15\textwidth]{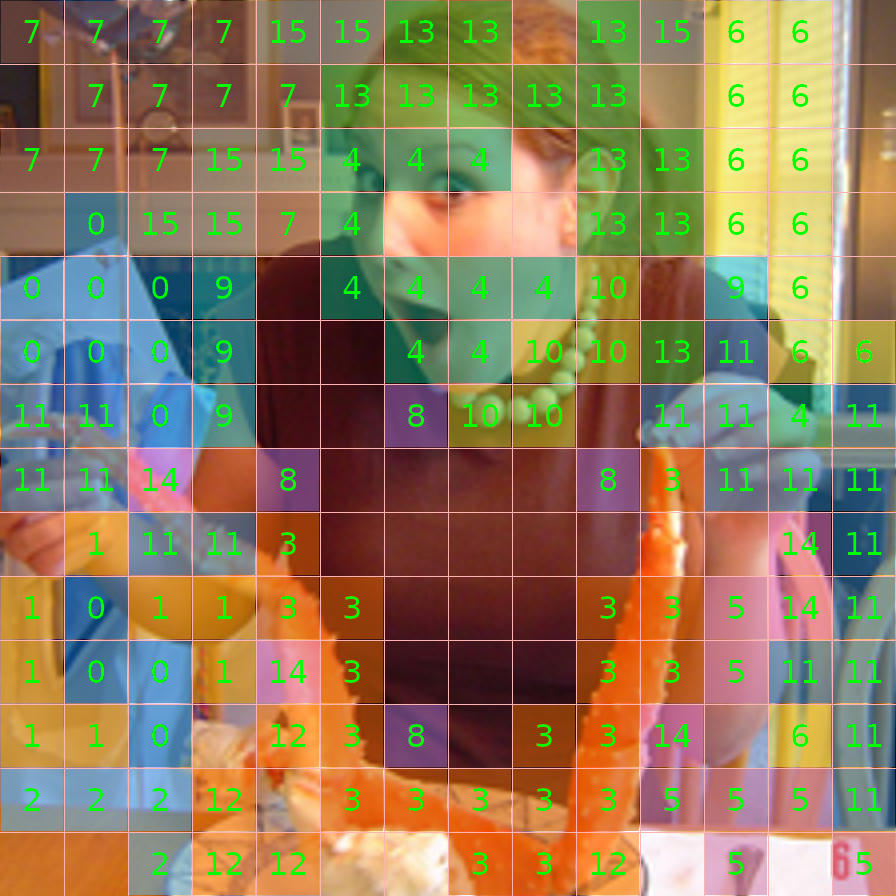} &
\includegraphics[width=0.15\textwidth]{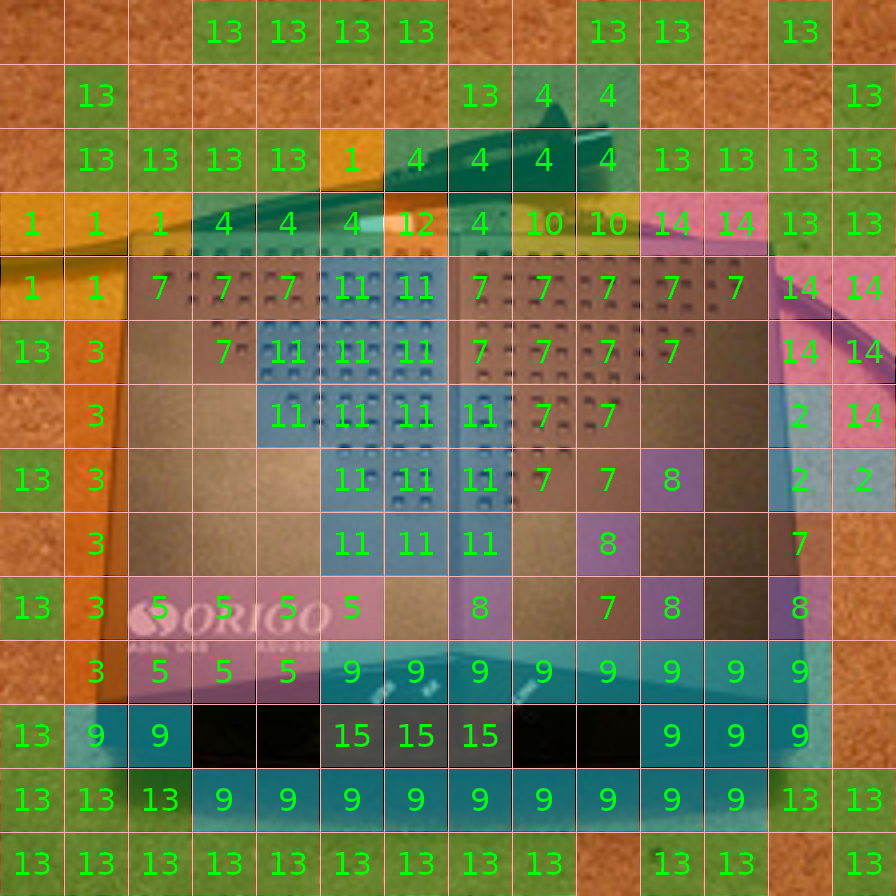} &
\includegraphics[width=0.15\textwidth]{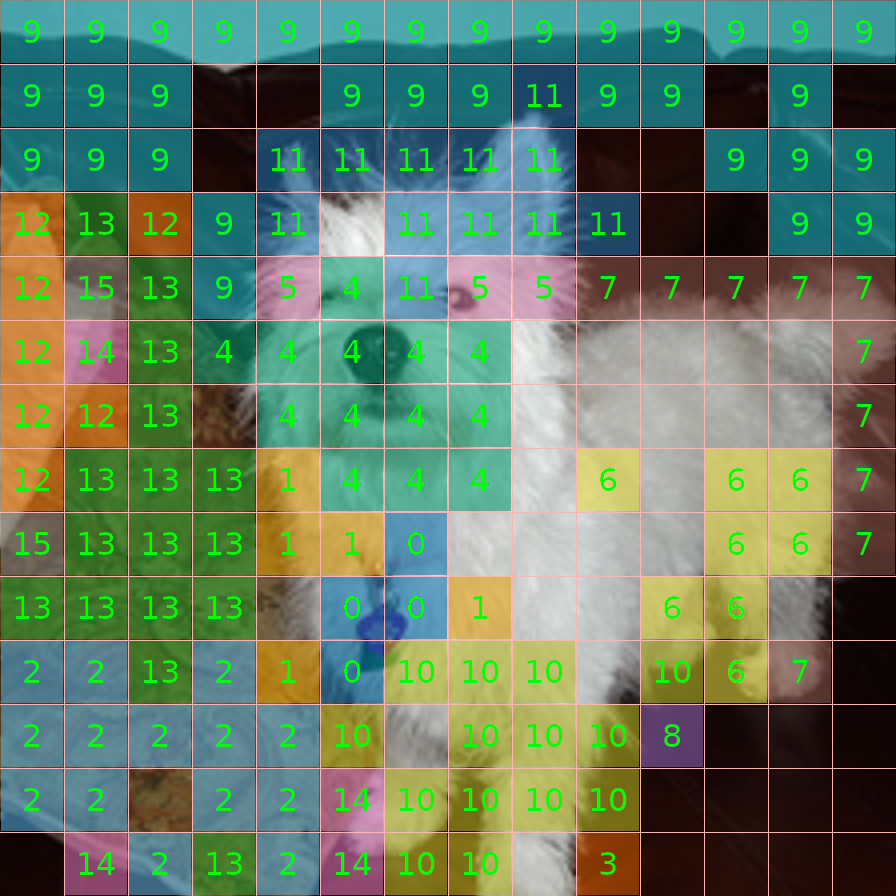}\\
\includegraphics[width=0.15\textwidth]{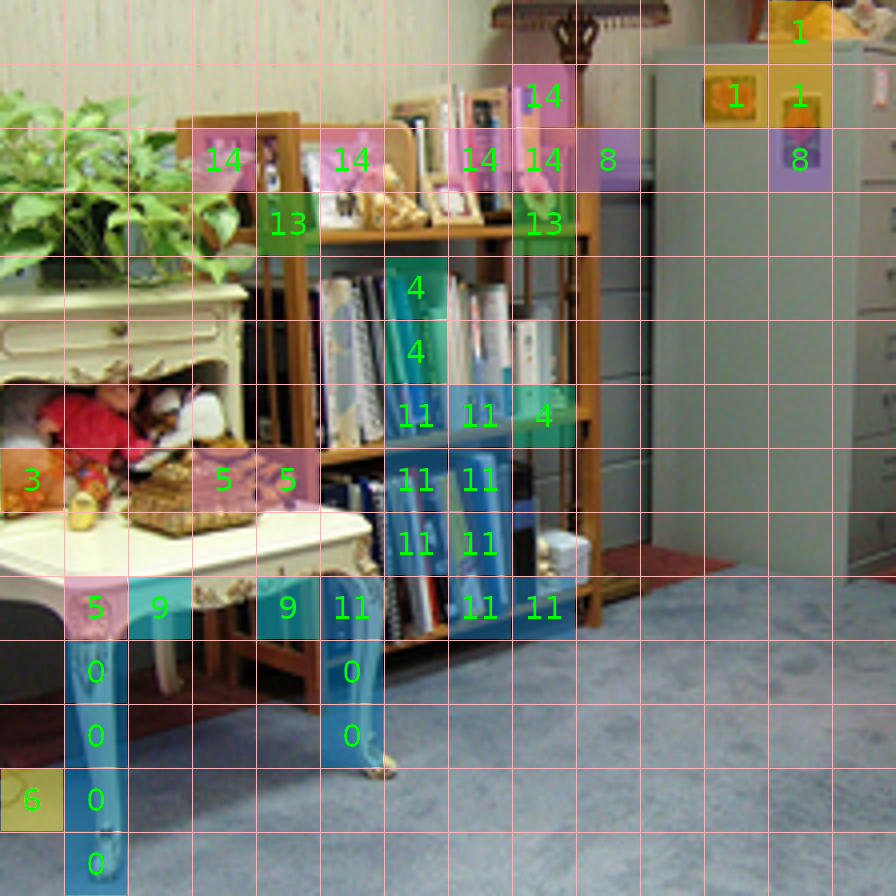} &
\includegraphics[width=0.15\textwidth]{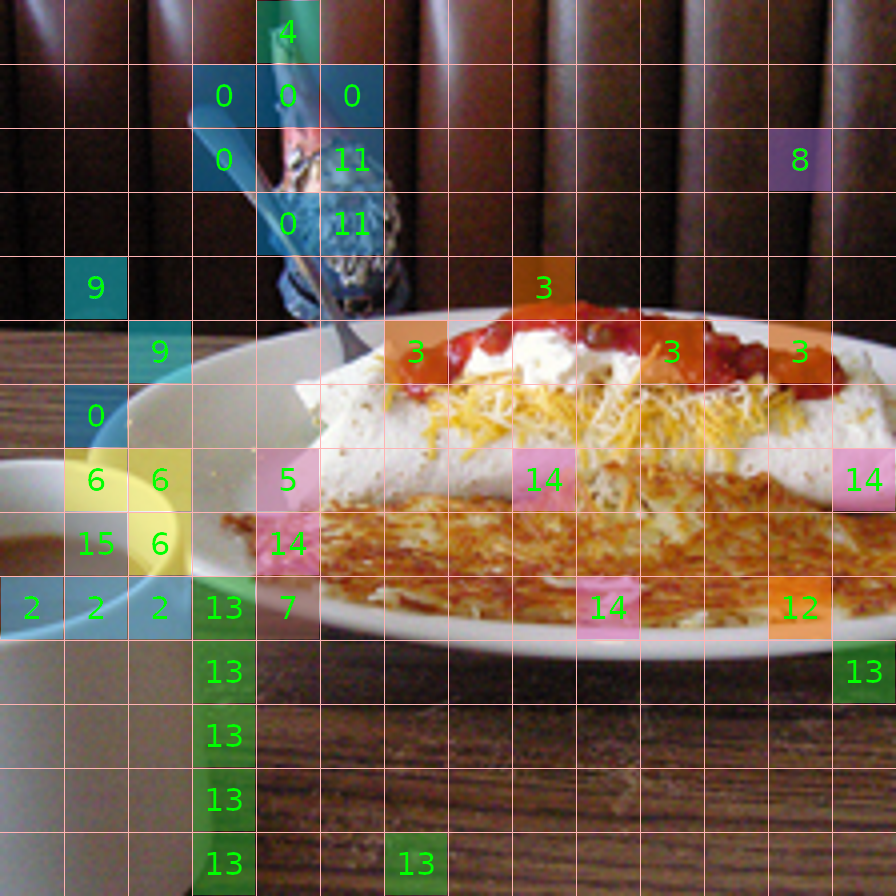} &
\includegraphics[width=0.15\textwidth]{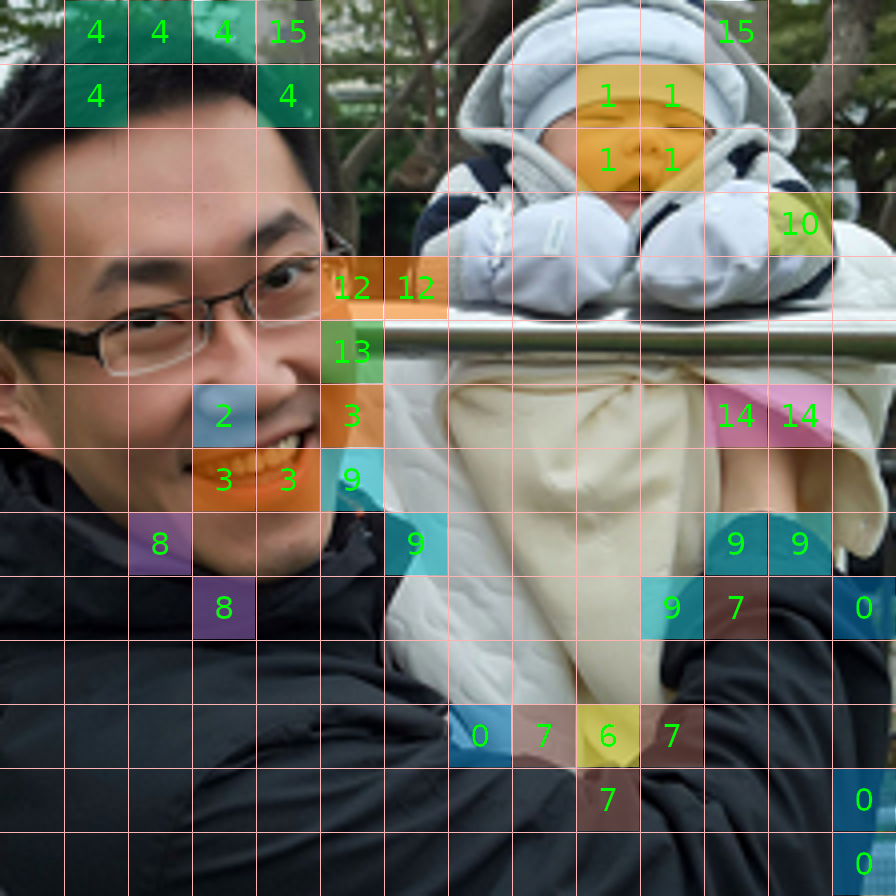} &
\includegraphics[width=0.15\textwidth]{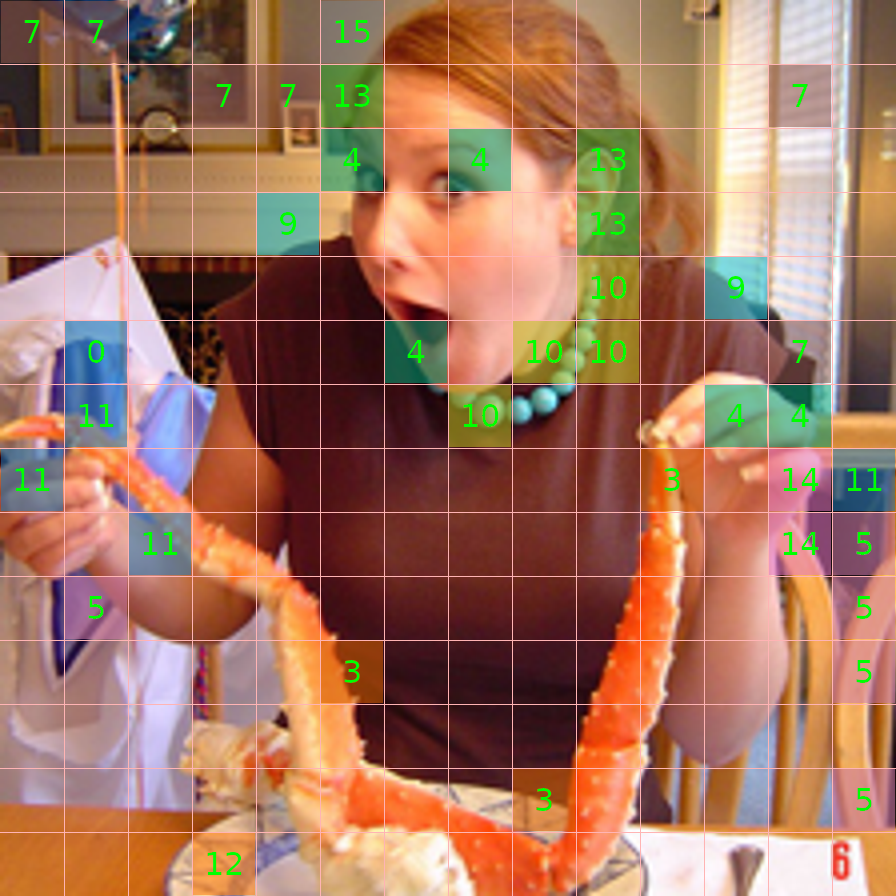} &
\includegraphics[width=0.15\textwidth]{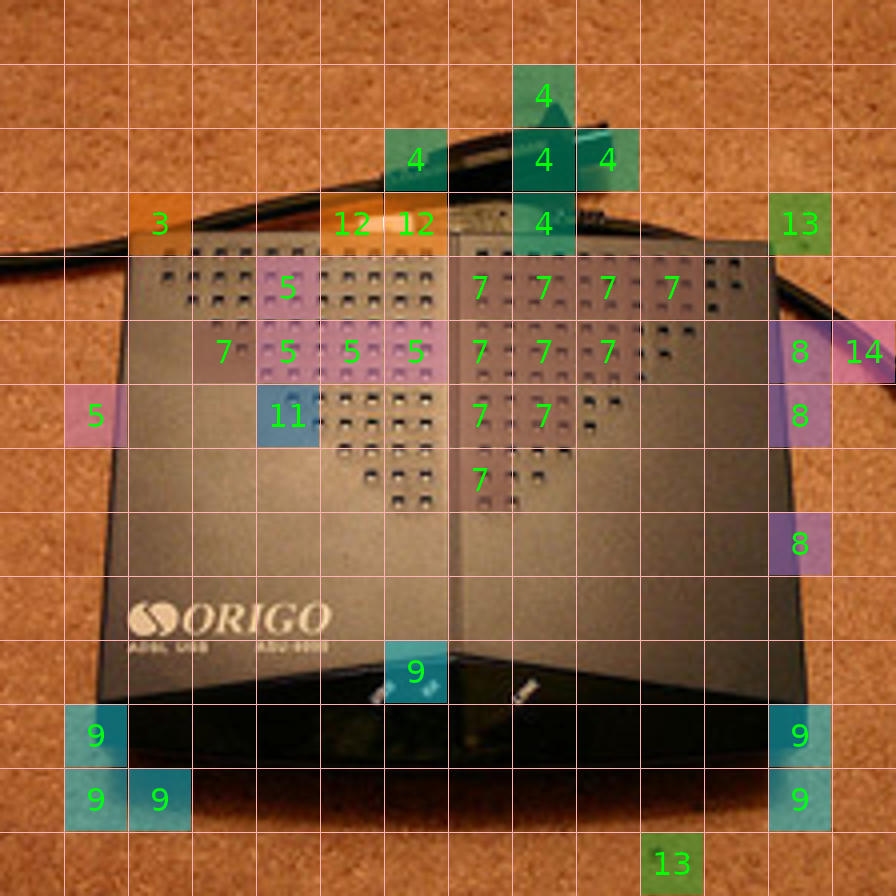} &
\includegraphics[width=0.15\textwidth]{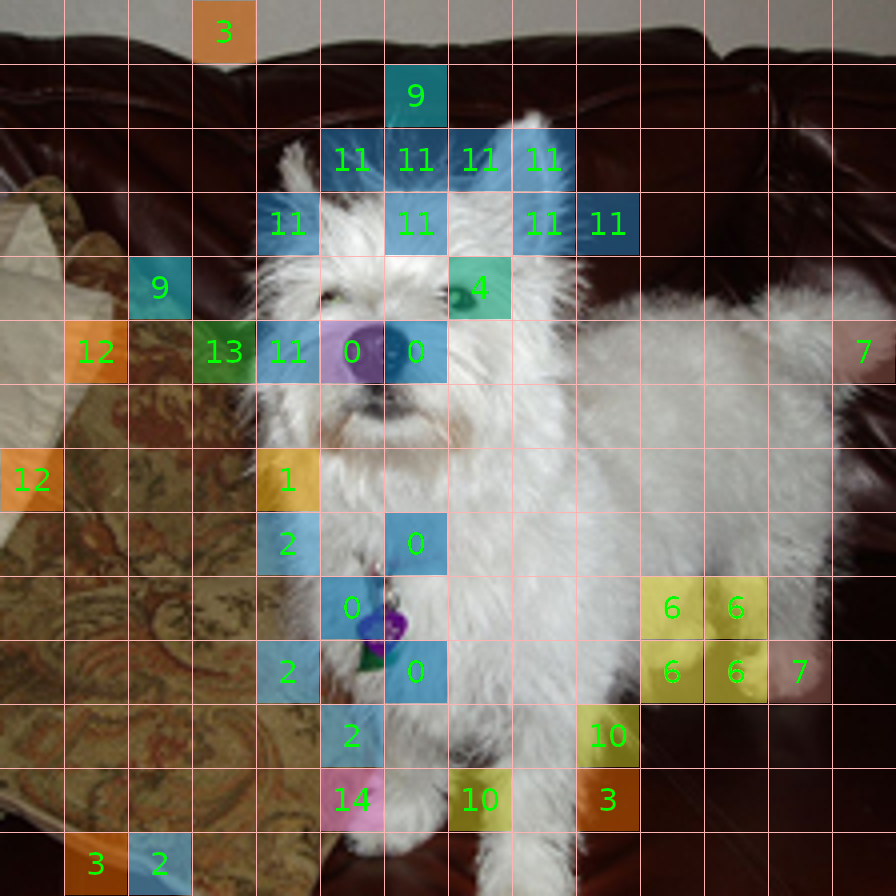}\\
\end{tabular}%
}
\endgroup
\caption{Patch-to-SMB assignments for tokens entering the first, second, and third compression layers of our A1A3A5 classification model, which use keep ratios of $0.75$, $0.25$, and $0.0$, respectively. The model's performance is reported in Table~\ref{tab:cls_SOTAcomparison}. Green numbers identify the SMB tokens assigned to reconstruct the corresponding patches, and patches assigned to the same SMB are masked with the same color. Unnumbered patches have been dropped by preceding layers. Examples are randomly selected from the ImageNet validation set.}
\label{fig:smb_vis_cls}
\end{figure*}

\section{Experiments}\label{sec:experiments}
% To evaluate whether the learned Summary Bank (SMB) compression learns highly condensed global semantics while preserving fine-grained visual details, and to assess the effectiveness, efficiency, and generality of the proposed method, we conduct experiments on image classification and vision-language understanding tasks.
\subsection{Image Classification}\label{sec:experimental-cls}
\subsubsection{Experimental Setup}\label{sec:experimental-cls-setup}
We evaluate image classification on ImageNet-1K~\cite{deng2009imagenet} using $224\times224$ inputs. Following prior work, we adopt DeiT-S~\cite{touvron2021training} as our primary backbone and insert compressive layers at designated positions within it. For example, A1A3A5 denotes inserting a compressive layer after each of Transformer blocks 1, 3, and 5.
%We also include RegNetY-16GF~\cite{radosavovic2020designing} as an alternative backbone to assess generality.
% TOLGA NOTE: I took out above statement because it is incorrect. We do not use a RegNet backbone, it is only used as a teacher.
% \paragraph{Implementation Details.}
We set the reconstruction loss weight to $\lambda=0.05$ for all experiments. When distillation is enabled, we apply a soft KL-divergence loss with a DeiT-S model trained from scratch as the teacher. The overall objective is
\begin{equation}
\mathcal{L}=\mathcal{L}_{\mathrm{task}}+\lambda\,\mathcal{L}_{\mathrm{rec}}, \quad
\mathcal{L}_{\mathrm{task}}=(1-\lambda_{\mathrm{kd}})\,\mathcal{L}_{\mathrm{ce}}+\lambda_{\mathrm{kd}}\,\mathcal{L}_{\mathrm{distill}},
\end{equation}
where we set $\lambda_{\mathrm{kd}}=0.5$ and use a temperature of $1.0$ for the distillation loss. We optimize all models with AdamW using a learning rate of $10^{-3}$, weight decay of $0.05$, and $\beta=(0.9,0.999)$, together with cosine learning-rate decay to $10^{-5}$. Training uses a batch size of 1024, and validation uses a batch size of 2048 on a single H200 GPU with 141GB memory. We report Top-1 accuracy.

% We train baseline models for 300 epochs with 5 warmup epochs, and compressive models for 500 epochs with 20 warmup epochs. 

\subsubsection{Main Results}\label{sec:experimental-cls-result}

Table~\ref{tab:cls_SOTAcomparison} shows that our method achieves a stronger efficiency--accuracy trade-off than prior approaches. We obtain 80.4\% Top-1 accuracy while reducing GMACs by 38\% relative to DeiT-S; even under a more aggressive 48\% reduction, our model still matches the backbone accuracy. We outperform TokenPooling, which achieves 100.2\% accuracy but reduces FLOPs by only 3.7\% relative to DeiT-S. In addition, we surpass methods that rely on importance-based pruning, such as Dynamic ViT~\cite{rao2021dynamicvit}, A-ViT~\cite{yin2022avit}, and ATS~\cite{fayyaz2022ats}, suggesting that hard token dropping based solely on saliency can be destructive when the removed tokens might need to support deeper semantic abstraction. Meanwhile, we achieve a better accuracy trade-off than EViT~\cite{liang2022evit}, Evo-ViT~\cite{xu2022evovit}, PPT~\cite{wu2023ppt}, and ToMe~\cite{bolya2023tome}, which preserve information through token merging or pooling. These results indicate that SMB can more effectively absorb visual information into a compact global representation.

\begin{table}[t]
\centering
\setlength{\belowcaptionskip}{6pt}
\caption{Comparison of computational cost and accuracy on ImageNet. The percentage in GMACs indicates the reduction computation relative to the backbone, and the percentage in accuracy indicates the retained performance relative to the backbone accuracy. The configurations ``0.75,0.25,0'' denote the keep ratios of the three compressive layers, corresponding to $\rho_{1}, \rho_{2}, \rho_{3}$ in Section~\ref{sec:method-recon_loss}. Our$_{distill}$ indicates that a DeiT-S teacher trained from scratch is used for distillation.}
\label{tab:cls_SOTAcomparison}
\fontsize{8.5pt}{10pt}\selectfont
\setlength{\tabcolsep}{12pt}
\renewcommand{\arraystretch}{0.92}
\setlength{\aboverulesep}{1pt}
\setlength{\belowrulesep}{1pt}
\begin{tabular}{lll}
\toprule
Model & GMACs & Accuracy \\
\midrule
DeiT-S(Base)  & 4.61\,(\text{100\%}) & 79.8\,(\text{100\%})\\
\midrule
A-ViT~\cite{yin2022avit}(CVPR'22)   & $3.6\,(\downarrow 22\%)$ & $78.6\,(\text{98.5\%})$ \\
STViT~\cite{chang2023making}(CVPR'23)   & $3.16\,(\downarrow31\%)$ & $80.6\,(\text{101.0\%})$ \\
EViT~\cite{liang2022evit}(ICLR'22)   & $3.0\,(\downarrow35\%)$ & $79.5\,(\text{99.6\%})$ \\
Evo-ViT~\cite{xu2022evovit}(AAAI'22)    & $3.0\,(\downarrow35\%)$ & $79.4\,(\text{99.5\%})$ \\
Dynamic ViT~\cite{rao2021dynamicvit}(NeurIPS'21)   & $2.9\,(\downarrow37\%)$ & $79.3\,(\text{99.4\%})$ \\
ATS~\cite{fayyaz2022ats}(ECCV‘22)   & $2.9\,(\downarrow37\%)$ & $79.7\,(\text{99.9\%})$ \\
PPT~\cite{wu2023ppt}(CoRR'22)   & $2.9\,(\downarrow37\%)$ & $79.8\,(\text{100.0\%})$ \\
ToMe~\cite{bolya2023tome}(ICLR'23)    & $2.7\,(\downarrow37\%)$ & $79.4\,(\text{99.5\%})$ \\
ReConV(Ours$_{\mathrm{distill}}$-A1A3A5[0.75,0.25,0.25])   & $2.88\,(\downarrow38\%)$ & $80.40\,(\text{100.8\%})$ \\
ReConV(Ours$_{\mathrm{distill}}$-A1A3A5[0.5,0.14,0.0])  & \textbf{$2.42\,(\downarrow48\%)$} & \textbf{$79.90\,(100.1\%)$} \\
\bottomrule
\end{tabular}
\end{table}

\paragraph{SMB Utilization and Semantic Partitioning}

As shown in Fig.~\ref{fig:smb_vis_cls}, patch-to-SMB assignments form spatially contiguous, semantically related regions despite receiving no spatial or semantic supervision. This suggests that the SMBs do not merely aggregate patches according to low-level proximity, but naturally develop specialization for recurring visual structures. Figure~\ref{fig:smb_vis_cls} also shows that the residual-norm dropping policy preferentially removes tokens from slowly varying regions, while retaining tokens near object boundaries and other high-frequency details. This is consistent with our design, which removes redundant structure while preserving visually informative content. We provide quantitative ablations in Appendix~\ref{cls_ablations}. Comparing SMB utilization maps from single- and three-layer compression models reveals more pronounced aggregation in the latter; similar behavior also emerges in multimodal tasks(see Appendix~\ref{sec:apdx-mllm-smb-vis}).

We analyze the SMB usage distribution over the full validation set and find that the SMBs are used approximately uniformly. For the A1A3A5 model, the corresponding entropies for the three compressive layers are 2.7030, 2.7531, and 2.7418 nats. These results indicate that the model makes effective use of the available SMB budget. For each SMB, we aggregate all tokens assigned to it across the validation set and construct a class distribution using the labels of their source images. For the 16-SMB model, the entropy values range from 6.1162 to 6.8889 nats, with a mean of 6.8219, close to the maximum of 6.9078 nats for 1000 classes. This high entropy indicates that no SMB is dominated by a small subset of classes. Together with the coherent assignment maps, this suggests that SMBs capture class-agnostic mid-level visual structure rather than acting as class-specific experts.

\subsection{Multi-Turn Multimodal Understanding}\label{sec:experimental-mllm}

\subsubsection{Experimental Setup}\label{sec:experimental-mllm-setup}

We instantiate ReConV on LLaVA-1.5~\cite{liu2024improved}, using CLIP-ViT-L-336px as the frozen vision encoder and Vicuna-13B as the LLM backbone, connected by an MLP projector. Two lightweight compressive layers are inserted between the vision encoder and projector, and the compressor and projector are trained from scratch. During pretraining, we optimize $\mathcal{L}=\mathcal{L}{\mathrm{task}}+\lambda\mathcal{L}{\mathrm{rec}}$ with $\lambda=0.01$. During instruction tuning, the projector is frozen and the LLM is updated with LoRA, while $\lambda$ is reduced to $0.001$ to preserve the learned compact representation during task adaptation. We use the 558K image-text alignment dataset and 665K multimodal instruction dataset released with LLaVA-1.5 for the two stages, respectively. Details are provided in Appendix~\ref{appendix:MLLM}. All training runs use $4\times$ H200 GPUs with 141 GB each, and evaluation uses $4\times$ L40S GPUs with 48 GB each.

\subsubsection{Main Results}\label{sec:experimental-mllm-result}
\paragraph{Evaluation protocol.}\label{sec:result-MLLM-mtQA}
Multi-turn dialogue is a natural setting for image-level compression because multiple questions share the same image. ReConV compresses each image once and reuses its visual representation across turns, amortizing visual encoding and compression costs. In contrast, prompt-conditioned selection depends on the current query and must be recomputed whenever the prompt changes. We evaluate this capability on MT-VQA-v2~\cite{goyal2017making} and ConvBench~\cite{liu2024convbench}. MT-VQA-v2 contains 3--14 questions per image, which we treat as successive independent queries, whereas ConvBench comprises native conversations with cross-turn dependencies. MetaCompress~\cite{wang2026rethinking} evaluates a similar setting using a manually constructed VQAv2 subset with fixed three-turn dialogues; however, neither the subset nor its implementation is publicly available, precluding a direct comparison. For ConvBench, we report direct-grading scores from Qwen3-VL-8B-Instruct on a 0--10 scale, using the same judge and prompt for all methods. 
% to ensure controlled within-table comparisons.

At a 95\% token reduction rate, ReConV achieves the best MT-VQA-v2 overall accuracy and ConvBench average among compressed methods, while providing $2.73\times$ and $1.51\times$ speedups, respectively. This favorable accuracy–efficiency trade-off under extreme compression validates the SMB's effectiveness in absorbing information from the complete visual sequence.

 % These scores support controlled within-table comparisons but are not directly comparable to results obtained under pairwise evaluation. 
 
\begin{figure}[t]
    \centering
    \includegraphics[width=\columnwidth,height=0.18\textheight]{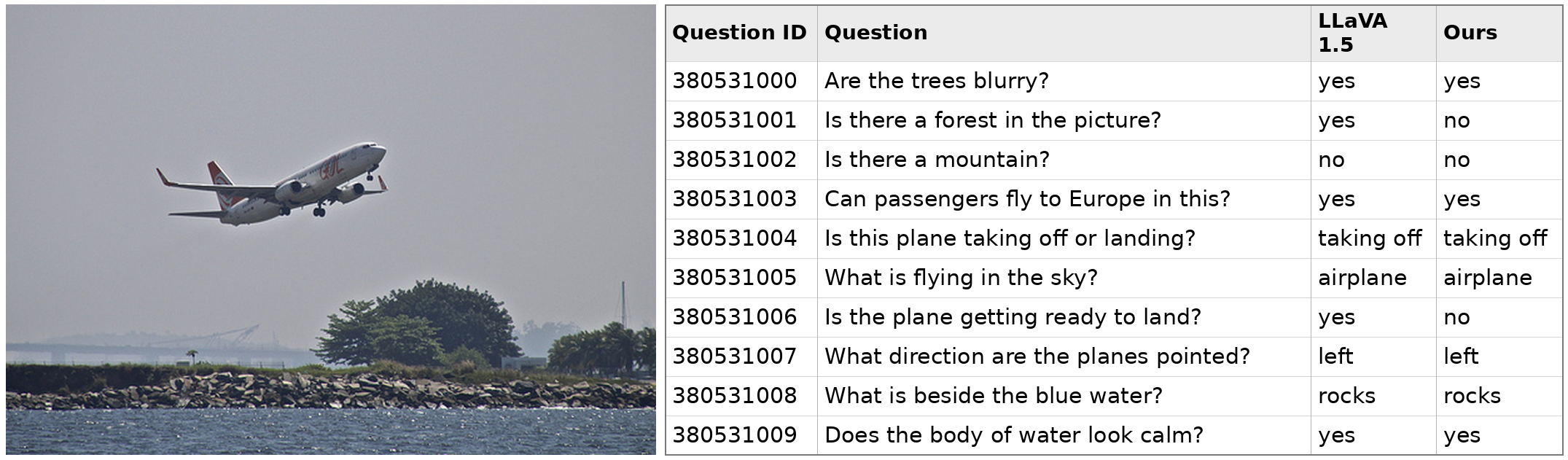}\\[-2pt]
    \includegraphics[width=\columnwidth,height=0.10\textheight]{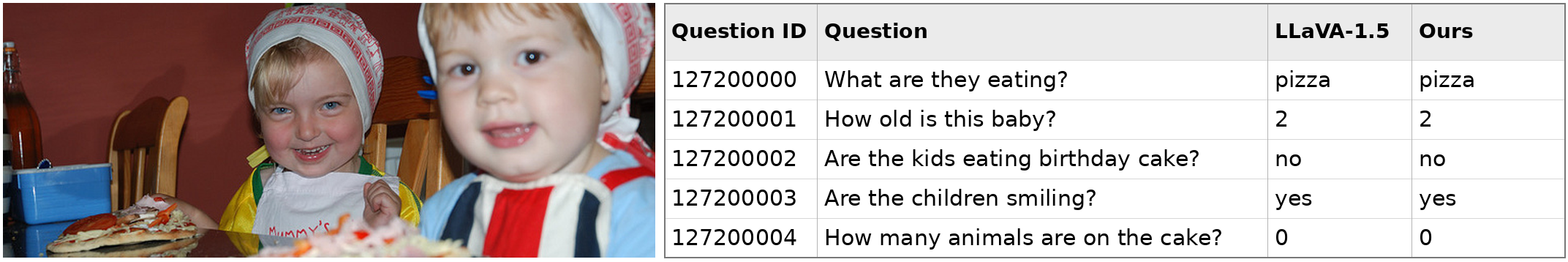}
    \caption{Examples from the MT-VQAv2 test split, where the number of questions associated with each image varies across samples. In the multi-turn setting, each image is encoded and optionally compressed only once, and the resulting visual embeddings are reused across all associated questions.}
    \label{fig:vqa_vis}
\end{figure}

\begin{table*}[t]
\centering
\caption{
Comparison on MT-VQA-v2 and ConvBench at a 95\% visual token reduction rate.
\emph{Amort.} denotes latency per question with image encoding amortized across questions, while \emph{Mean Total Prefill} denotes total prefill latency averaged across conversations.
}
\label{tab:multiturn_comparison}
\setlength{\tabcolsep}{2.6pt}
\renewcommand{\arraystretch}{1.00}
\scriptsize

\begin{tabular*}{\textwidth}{
    @{\extracolsep{\fill}}
    l
    *{12}{c}
    @{}
}
\toprule
\multirow{2}{*}{\textbf{Method}}
& \multicolumn{6}{c}{\textbf{MT-VQA-v2}}
& \multicolumn{6}{c}{\textbf{ConvBench}} \\
\cmidrule(lr){2-7}
\cmidrule(lr){8-13}
& \textbf{Yes/No}
& \textbf{Number}
& \textbf{Other}
& \textbf{Overall}
& \shortstack{\textbf{Amort.}\\\textbf{(ms/q)} $\downarrow$}
& \shortstack{\textbf{Speedup}\\\textbf{($\times$)} $\uparrow$}
& $S_1$
& $S_2$
& $S_3$
& \textbf{Average}
& \shortstack{\textbf{Mean Total}\\\textbf{Prefill (s)} $\downarrow$}
& \shortstack{\textbf{Speedup}\\\textbf{($\times$)} $\uparrow$} \\
\midrule
Uncompressed
& 93.28 & 62.22 & 72.22 & 79.99
& 314.50 & 1.00
& 3.47 & 4.33 & 4.68 & 4.16
& 361.05 & 1.00 \\
Random Pruning
& 87.19 & 47.43 & 59.97 & 69.76
&114.91  & 2.74
& 3.29 & 3.29 & 2.60 & 3.06
& 242.06 & 1.49 \\
Spatial Pooling
& 80.41 & 38.60 & 50.58 & 61.51
& 114.32 & 2.75
& 3.12 & 3.12 & 2.77 & 3.00
& 240.92 & 1.50 \\
FastV\citenumonly{chen2024fastv}
& 80.56 & 40.75 & 54.15 & 63.51
& 192.40 & 1.63
& 3.81 & 3.64 & 3.47 & 3.64
&739.47  & 0.49 \\
PDrop\citenumonly{xing2024pyramiddrop}
&75.71& 30.56 & 42.05 & 54.61 
& 119.30 & 2.64
&2.60  & 2.95 & 2.60 & 2.72
& 237.76 & 1.52 \\
PruMerge+\citenumonly{shang2025llavaprumerge}
& 83.38 & 45.67 & 59.94 & 67.97
& 117.17 & 2.68
& {3.72} & {3.80} & 3.46 & {3.66}
& {239.79} & {1.51} \\
SparseVLM+\citenumonly{zhang2024sparsevlm}
&79.55  & 32.78 & 48.89 & 59.69
& 126.79 & 2.48
&2.08  & 3.99 & 1.91 & 2.66
& 238.65 & 1.51 \\
ReConV(Ours)
% & 87.64 & 51.76 & 62.94 & \textbf{71.85}
& {86.85} & {50.82}& {61.92} & {70.93}
& 115.36 & 2.73
& 3.12 & {3.81} & {4.16}
& {3.70} & {239.81}  & {1.51} \\
\bottomrule
\end{tabular*}
\end{table*}

\subsection{Ablation Study}\label{sec:ablation}
% \subsubsection{Analysis of the compute--accuracy trade-off}\label{sec:ablation-cls-tradeoff}
\begin{figure*}[t]
\centering
\begingroup
\setlength{\tabcolsep}{4pt} 
\begin{tabular}{cc}
\includegraphics[width=0.485\textwidth]{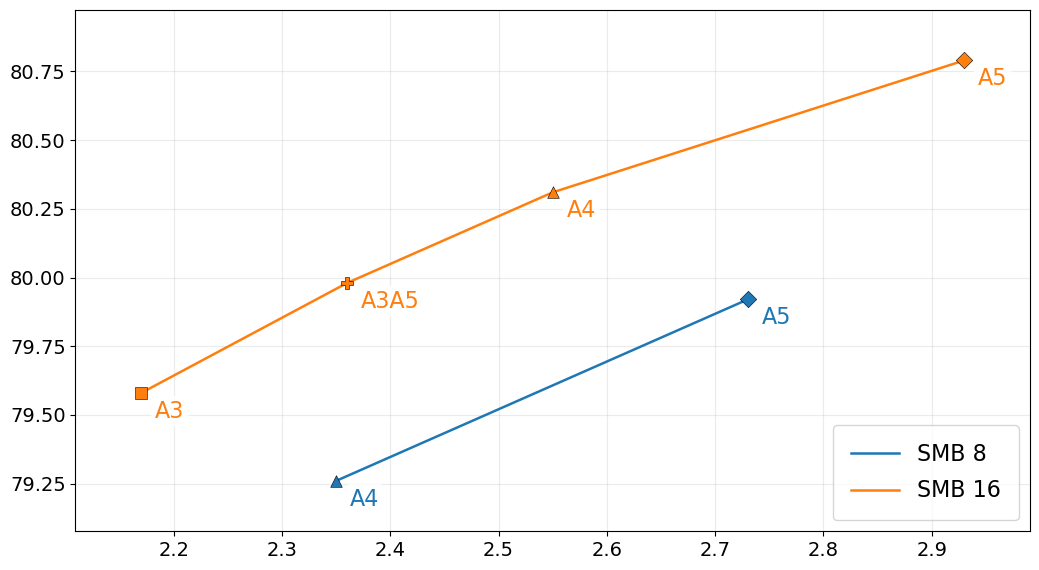} &
\includegraphics[width=0.485\textwidth]{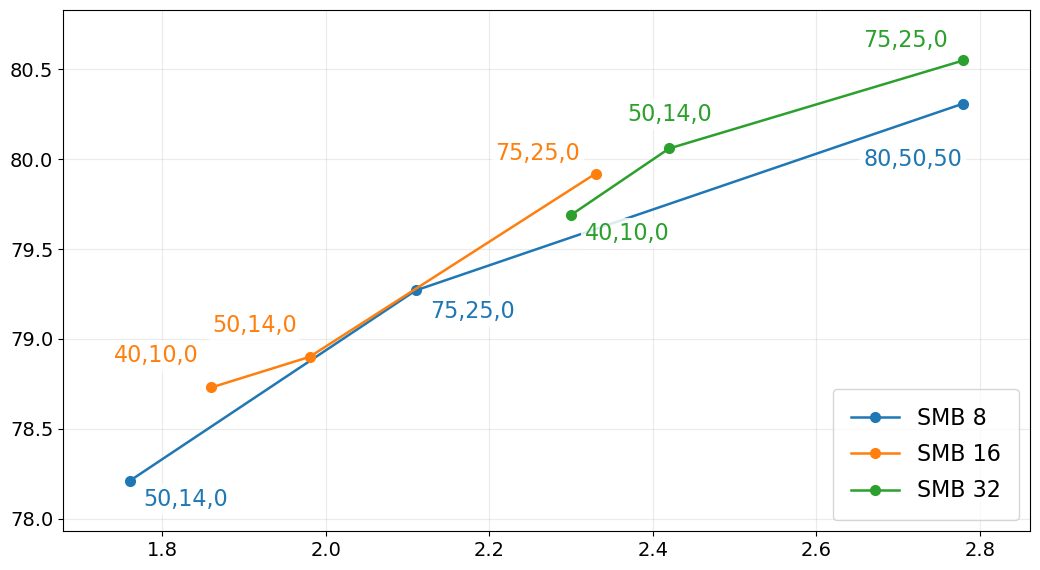}
\end{tabular}
\endgroup
\label{fig:cls_tradeoff}
\caption{ImageNet1K top-1 accuracy percentage vs computation cost (GMACs) tradeoff. Left: Models with 1-2 compressive layers with varying number of SMBs and layer placement. Right: Models with 3 compressive layers (A1A3A5) with varying number of SMBs and keep ratios. }
\end{figure*}

We analyze the accuracy--compute trade-off by varying compressive-layer placement, keep ratios, and the number of SMBs. Figure~\ref{fig:cls_tradeoff} (Left) shows that, for both 8 and 16 SMBs, earlier compression reduces computation but degrades accuracy, indicating that aggressive reduction is more harmful in shallow layers. The two-layer A3A5 model with keep ratios of $0.19$ and $0.0$ lies on the same approximately linear frontier as the A3 and A5 configurations, while reducing the SMB count from 16 to 8 produces a less favorable trade-off. Additional results are reported in Appendix~\ref{full_cls_results}.

Figure~\ref{fig:cls_tradeoff} (Right) further examines different keep-ratio schedules for models with 8, 16, and 32 SMBs, ranging from conservative schedules with less reduction in early layers to aggressive schedules with stronger early-layer compression. The annotated values indicate the keep ratios applied at each compressive layer. Across these configurations, the models exhibit a consistent and approximately linear accuracy--compute trend, suggesting that the proposed framework is robust to different configuration choices. Notably, at a similar computational cost, the 32-SMB model with a $0.75$, $0.25$, $0.0$ keep-ratio schedule outperforms the 8-SMB model with a $0.80$, $0.50$, $0.50$ schedule, indicating that a larger summary bank provides a more effective compression basis for classification.
% Figure~\ref{fig} (Right) compares keep-ratio schedules across 8, 16, and 32 SMBs. The models follow a consistent, approximately linear accuracy--compute trend, indicating robustness to configuration choices. At similar computational cost, the 32-SMB model with a $0.75$, $0.25$, $0.0$ schedule outperforms the 8-SMB model with a $0.80$, $0.50$, $0.50$ schedule, showing that a larger summary bank provides a more effective compression basis.

\begin{table*}[t]
\centering
\caption{
Ablation study of the compression design on multimodal understanding.
\emph{Inter-layer repr.} and \emph{LLM input} specify whether original or residual tokens are propagated between compression layers and supplied to the LLM, respectively.
\(K\) denotes the total token budget, including retained and SMB tokens.
\emph{MME-P} and \emph{MME-C} denote the total perception and cognition scores.
}
\label{tab:method_ablation}
\setlength{\tabcolsep}{3.5pt}
\renewcommand{\arraystretch}{1.00}
\scriptsize
\begin{tabular*}{0.98\textwidth}{
    @{\extracolsep{\fill}}
    l
    *{9}{c}
    @{}
}
\toprule
\textbf{Ablation}
& \shortstack{\textbf{Inter-layer}\\\textbf{repr.}}
& \shortstack{\textbf{LLM}\\\textbf{input}}
& \shortstack{\textbf{Position}\\\textbf{reinjection}}
& \shortstack{\textbf{Total}\\\textbf{tokens}}
& \shortstack{\textbf{\# Retained}\\\textbf{tokens}}
& \shortstack{\textbf{\# SMB}\\\textbf{tokens}}
& \shortstack{\textbf{\# Compression}\\\textbf{layers}}
& \shortstack{\textbf{MME-P}\\\textbf{Score}}
& \shortstack{\textbf{MME-C}\\\textbf{Score}} \\
\midrule

\textcolor{black!40}{LLaVA-1.5-13B (Base)}
& \multicolumn{7}{c}{\textcolor{black!40}{\textit{Uncompressed (576 tokens)}}}
& \textcolor{black!40}{1531}
& \textcolor{black!40}{295} \\
\midrule

\multirow{4}{*}{\shortstack[l]{Representation\\propagation}}
& Original & Original & \cmark &127& 119 & 8 & 2 & 1242 & 273 \\
& Original & Residual & \cmark &127& 119 & 8 & 2 & 1370 & 285 \\
& Residual & Original & \cmark &127& 119 & 8 & 2 & 1405 & 305 \\
& Residual & Residual & \cmark &127& 119 & 8 & 2 & 1457 & 291 \\
\midrule

\multirow{2}{*}{\shortstack[l]{Positional\\information}}
& Residual & Residual & \xmark &127 & 119 & 8 & 2 & 1337 & 285 \\
& Residual & Residual & \cmark &127 & 119 & 8 & 2 & 1457 & 291\\
\midrule

\multirow{3}{*}{\shortstack[l]{Allocation\\($K=32$)}}
& Residual & Residual & \cmark
& 32 & 1  & 31 & 2 & 1337 & 291 \\
& Residual & Residual & \cmark
& 32 & 16 & 16 & 2 & 1304 & 293 \\
& Residual & Residual & \cmark
& 32 & 24 & 8  & 2 & 1352 & 288 \\
\midrule

\multirow{2}{*}{\shortstack[l]{Allocation\\($K=64$)}}
& Residual & Residual & \cmark
& 64 & 32 & 32 & 2 & 1327 & 321 \\
& Residual & Residual & \cmark
& 64 & 56 & 8  & 2 & 1404 & 321 \\
\midrule

\multirow{4}{*}{\shortstack[l]{Token budget\\(8 SMBs)}}
& Residual & Residual & \cmark
& 127 & 119 & 8 & 2 & 1457 & 291 \\
& Residual & Residual & \cmark
& 144 & 136 & 8 & 2 & 1423 & 286 \\
& Residual & Residual & \cmark
& 192 & 184 & 8 & 2 & 1457 & 282 \\
& Residual & Residual & \cmark
& 263 & 255 & 8 & 2 & 1519 & 289 \\
\midrule

\multirow{3}{*}{Compression depth}
& Residual & Residual & \cmark & 32 & 24 & 8 & 2 & 1352 & 288 \\
& Residual & Residual & \cmark & 32 & 24 & 8 & 3 & 1325 & 312 \\
\bottomrule
\end{tabular*}
\end{table*}

% \paragraph{Ablation analysis.}
On MME~\cite{fu2023mme}, we ablate compression depth, positional reinjection, token allocation, and representation type. Re-adding positional embeddings before each compression layer improves MME-P from 1337 to 1457 and MME-C from 285 to 291. Under a fixed budget, allocating more tokens to explicit retention generally performs better: at \(K=64\), changing the retained/SMB allocation from \(32+32\) to \(56+8\) raises MME-P from 1327 to 1404 while preserving MME-C. Using residuals for both inter-layer propagation and LLM input achieves the best overall performance, supporting their role in propagating information not absorbed by the SMB. Across budgets, retaining approximately 50\% of the visual tokens reduces MME-P by only 0.8\%, while a 95\% reduction incurs an 11.7\% drop, demonstrating a favorable performance--efficiency trade-off.

\section{Conclusion}\label{sec:conclusion}
In this paper, we introduced a reconstruction-guided framework for visual token compression. By compressing token sequences into a compact set of learnable Summary Bank tokens and using reconstruction residuals to guide token retention, our method enables hard token dropping without hard semantic deletion. Experiments on image classification and multimodal understanding demonstrate favorable performance--efficiency trade-offs across tasks.

% \clearpage
\bibliography{iclr2027_conference}
\bibliographystyle{iclr2027_conference}

\section*{AI Use Statement}
Generative AI tools were used solely to improve the grammar, clarity, and readability of author-written text. They were not used to develop the research methodology, conduct experiments, analyze results, or formulate scientific claims. All AI-assisted edits were reviewed and revised by the authors, who take full responsibility for the final content of this paper.

\clearpage

\appendix
\section{Appendix}
\subsection{Implementation}
\label{appendix:1}
Algorithm~\ref{alg:compressive_layer_pesudocode} summarizes one compressive layer. The layer first updates a small summary bank by asymmetric cross-attention, then reconstructs each token from its routed summary slot and positional embedding, and finally retains only the tokens with the largest reconstruction residuals.

\begin{algorithm}[t]
\caption{A Reconstruction-Guided Compressive Layer}
\label{alg:compressive_layer_pesudocode}
\begin{algorithmic}[1]
\Require tokens $\mathbf{X}\in\mathbb{R}^{N\times D}$, summary bank $\mathbf{S}\in\mathbb{R}^{M\times D}$, positions $\mathbf{P}$, keep ratio $\rho$, loss type $\tau\in\{\texttt{L2},\texttt{NCE}\}$
\Ensure updated bank $\mathbf{S}'$, kept tokens $\mathbf{X}_{\mathrm{keep}}$, loss $\mathcal{L}_{\mathrm{rec}}$

\State $(\Delta\mathbf{S}, \mathbf{L}) \gets \mathrm{CrossAttnWithLogits}(\mathrm{LN}(\mathbf{S}), \mathrm{LN}(\mathbf{X}), \mathrm{LN}(\mathbf{X}))$
\State $\mathbf{S}' \gets \mathbf{S} + \Delta\mathbf{S}$
\State $\mathbf{S}' \gets \mathbf{S}' + \mathrm{MLP}_{\mathrm{smb}}(\mathrm{LN}(\mathbf{S}'))$
\If{SMB-only MHSA is enabled}
    \State $\mathbf{S}' \gets \mathbf{S}' + \mathrm{MHSA}_{\mathrm{smb}}(\mathrm{LN}(\mathbf{S}'))$
\EndIf

\For{$i=1,\dots,N$}
    \State $m_i^\star \gets \arg\max_m \mathbf{L}_{m,i}$
    \State $\hat{\mathbf{x}}_i \gets g_\theta([\phi_s(\mathbf{S}'_{m_i^\star}); \phi_p(\mathbf{P}_i)])$
    \If{$\tau=\texttt{L2}$}
        \State $\mathbf{r}_i \gets \mathbf{x}_i-\hat{\mathbf{x}}_i$
    \Else
        \State $\mathbf{u}_i \gets \hat{\mathbf{x}}_i/\|\hat{\mathbf{x}}_i\|_2$
        \State $\mathbf{r}_i \gets \mathbf{x}_i-(\mathbf{x}_i^\top\mathbf{u}_i)\mathbf{u}_i$
    \EndIf
    \State $s_i \gets \|\mathbf{r}_i\|$
\EndFor

\State $k \gets \max(1,\mathrm{round}(N\rho))$
\State $\mathcal{I}_{\mathrm{keep}} \gets \mathrm{TopKIndices}(\{s_i\}_{i=1}^N,k)$
\State $\mathbf{X}_{\mathrm{keep}} \gets \{\mathbf{r}_i : i\in\mathcal{I}_{\mathrm{keep}}\}$

\State $\Omega \gets \{1,\dots,N\}\setminus\mathcal{I}_{\mathrm{keep}}$ if dropped-only loss else $\{1,\dots,N\}$

\If{$\tau=\texttt{L2}$}
    \State $\mathcal{L}_{\mathrm{rec}} \gets \frac{1}{|\Omega|}\sum_{i\in\Omega}\frac{\|\mathbf{x}_i-\hat{\mathbf{x}}_i\|_2^2}{D}$
\Else
    \State sample anchors from $\Omega$
    \State form positive logits from $(\mathbf{x}_i,\hat{\mathbf{x}}_i)$ and negatives from same-SMB wrong positions and other images
    \State $\mathcal{L}_{\mathrm{rec}} \gets \mathrm{InfoNCE}(\text{logits})$
\EndIf

\State \Return $\mathbf{S}'$, $\mathbf{X}_{\mathrm{keep}}$, $\mathcal{L}_{\mathrm{rec}}$
\end{algorithmic}
\end{algorithm}

\section{Experiments}

\subsection{Classification Setup}
% The classification architecture is illustrated in Figure~\ref{fig:classification_architecture}. 
We follow the standard timm ImageNet training recipe, including stochastic depth with rate 0.1, RandAugment (rand-m9-mstd0.5-inc1), bicubic interpolation, color jitter of 0.4, random erasing with probability 0.25 in pixel mode, label smoothing of 0.1, mixup with $\alpha=0.8$, CutMix with $\alpha=1.0$, mixup probability 1.0, and mixup/CutMix switching probability 0.5. During evaluation, we use a crop ratio of 0.875 with bicubic interpolation. Optimization is performed with AdamW (learning rate $10^{-3}$, weight decay $0.05$, $\beta=(0.9, 0.999)$), cosine learning-rate decay to $10^{-5}$, gradient clipping at 1.0, automatic mixed precision, and no gradient accumulation. The vanilla models are trained for 300 epochs with 5 warmup epochs, while the compressive models are trained for 500 epochs with 20 warmup epochs. We also maintain an exponential moving average of the student weights during training (decay 0.99996) and use it for evaluation.

\subsubsection{Results}\label{full_cls_results}
We first train a vanilla DeiT-S-style model without distillation and obtain 80.61\% top-1 accuracy on ImageNet. We also train a DeiT-S-style model with RegNet-Y-160 as the teacher, which achieves 81.88\% top-1 accuracy. Based on these two backbones, we train two sets of compressive models: one distilled from the vanilla model and one distilled from the stronger RegNet-Y-160 teacher. Full results are given in Table~\ref{tab:cls_Results_apdx}. For the placement study, we evaluate a single compressive layer with a keep ratio of $0.0$. This setting retains only the SMB tokens and removes all patch tokens after the compressive layer, similar to TokenLearner~\cite{ryoo2021tokenlearner} and STViT~\cite{chang2023making}. 

\begin{table*}[t]
\centering
\caption{ImageNet1k top-1 accuracy and computational cost of the proposed compressive model under various settings. comp-f and comp-r refer to our model using compressive layers that output full vs residual tokens, respectively.}
\label{tab:cls_Results_apdx}
\footnotesize
\setlength{\tabcolsep}{2.2pt}
\renewcommand{\arraystretch}{1.12}
\resizebox{\textwidth}{!}{%
\begin{tabular}{l c c c c c c c c c}
\toprule
\textbf{Model} & \textbf{Heads} & \textbf{Dim} & \textbf{Layers} & \textbf{Teacher} & \textbf{SMB} & \textbf{Configuration} & \textbf{Keep ratios} & \textbf{GMACs} & \textbf{Accuracy} \\
\midrule
Vanilla & 6 & 384 & 12 & -- & -- & -- & -- & 4.61 & 80.61 \\
DeiT-S & 6 & 384 & 12 & regnet-y-160 & -- & -- & -- & 4.61 & 81.88 \\
\midrule
Comp-R & 6 & 384 & 12 & Vanilla & 16 & A1A3A5 & 0.75,0.25,0.25 & 2.43 & 79.61 \\
Comp-F & 6 & 384 & 12 & Vanilla & 16 & A1A3A5 & 0.75,0.25,0.25 & 2.43 & 79.64 \\
Comp-R & 6 & 384 & 12 & Vanilla & 16 & A1A3A5 & 0.75,0.25,0.00 & 2.33 & 79.75 \\
Comp-F & 6 & 384 & 12 & Vanilla & 32 & A1A3A5 & 0.75,0.25,0.25 & 2.88 & 80.44 \\
Comp-R & 6 & 384 & 12 & Vanilla & 32 & A1A3A5 & 0.50,0.14,0.00 & 2.42 & 79.79 \\
Comp-F & 6 & 384 & 12 & Vanilla & 32 & A1A3A5 & 0.50,0.14,0.00 & 2.42 & 79.90 \\
\midrule
Comp-F & 6 & 384 & 12 & regnet-y-160 & 16 & A3 & 0.00 & 2.17 & 79.58 \\
Comp-F & 6 & 384 & 12 & regnet-y-160 & 8 & A4 & 0.00 & 2.35 & 79.26 \\
Comp-F & 6 & 384 & 12 & regnet-y-160 & 16 & A4 & 0.00 & 2.55 & 80.31 \\
Comp-F & 6 & 384 & 12 & regnet-y-160 & 8 & A5 & 0.00 & 2.73 & 79.92 \\
Comp-F & 6 & 384 & 12 & regnet-y-160 & 16 & A5 & 0.00 & 2.93 & 80.79 \\
Comp-F & 6 & 384 & 12 & regnet-y-160 & 16 & A3A5 & 0.19,0.00 & 2.36 & 79.98 \\
Comp-F & 6 & 384 & 12 & regnet-y-160 & 8 & A1A3A5 & 0.80,0.50,0.50 & 2.78 & 80.31 \\
Comp-F & 6 & 384 & 12 & regnet-y-160 & 8 & A1A3A5 & 0.75,0.25,0.25 & 2.20 & 79.26 \\
Comp-F & 6 & 384 & 12 & regnet-y-160 & 8 & A1A3A5 & 0.75,0.25,0.00 & 2.11 & 79.27 \\
Comp-F & 6 & 384 & 12 & regnet-y-160 & 8 & A1A3A5 & 0.50,0.14,0.00 & 1.76 & 78.21 \\
Comp-F & 6 & 384 & 12 & regnet-y-160 & 16 & A1A3A5 & 0.75,0.25,0.25 & 2.43 & 79.99 \\
Comp-F & 6 & 384 & 12 & regnet-y-160 & 16 & A1A3A5 & 0.75,0.25,0.00 & 2.33 & 79.92 \\
Comp-F & 6 & 384 & 12 & regnet-y-160 & 16 & A1A3A5 & 0.50,0.14,0.00 & 1.98 & 78.90 \\
Comp-F & 6 & 384 & 12 & regnet-y-160 & 16 & A1A3A5 & 0.40,0.10,0.00 & 1.86 & 78.73 \\
Comp-F & 6 & 384 & 12 & regnet-y-160 & 32 & A1A3A5 & 0.75,0.25,0.00 & 2.78 & 80.55 \\
Comp-F & 6 & 384 & 12 & regnet-y-160 & 32 & A1A3A5 & 0.50,0.14,0.00 & 2.42 & 80.06 \\
Comp-F & 6 & 384 & 12 & regnet-y-160 & 32 & A1A3A5 & 0.40,0.10,0.00 & 2.30 & 79.69 \\
\bottomrule
\end{tabular}%
}
\end{table*}

\paragraph{SMB Partition}\label{sec:apdx-cls-smb-vis}
The color-coded masks exhibit semantic aggregation, suggesting that different SMB tokens tend to absorb distinct types of semantic information. This tendency is more pronounced in the A1A3A5 model than in the A5 model when comparing the tokens entering their first compression layers. This observation suggests that progressive, multi-layer compression may promote more semantically coherent aggregation than single-layer compression.

\begin{figure*}[t]
\centering
\begingroup
\setlength{\tabcolsep}{4pt} 
\begin{tabular}{cccccc}
\includegraphics[width=0.15\textwidth]{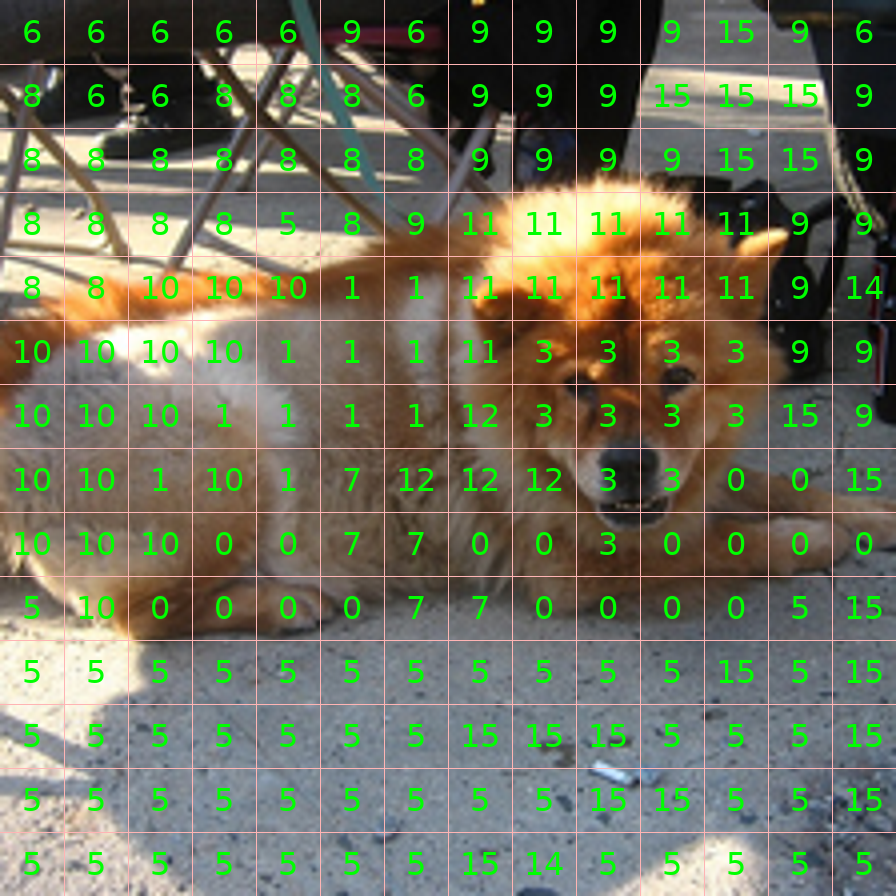} &
\includegraphics[width=0.15\textwidth]{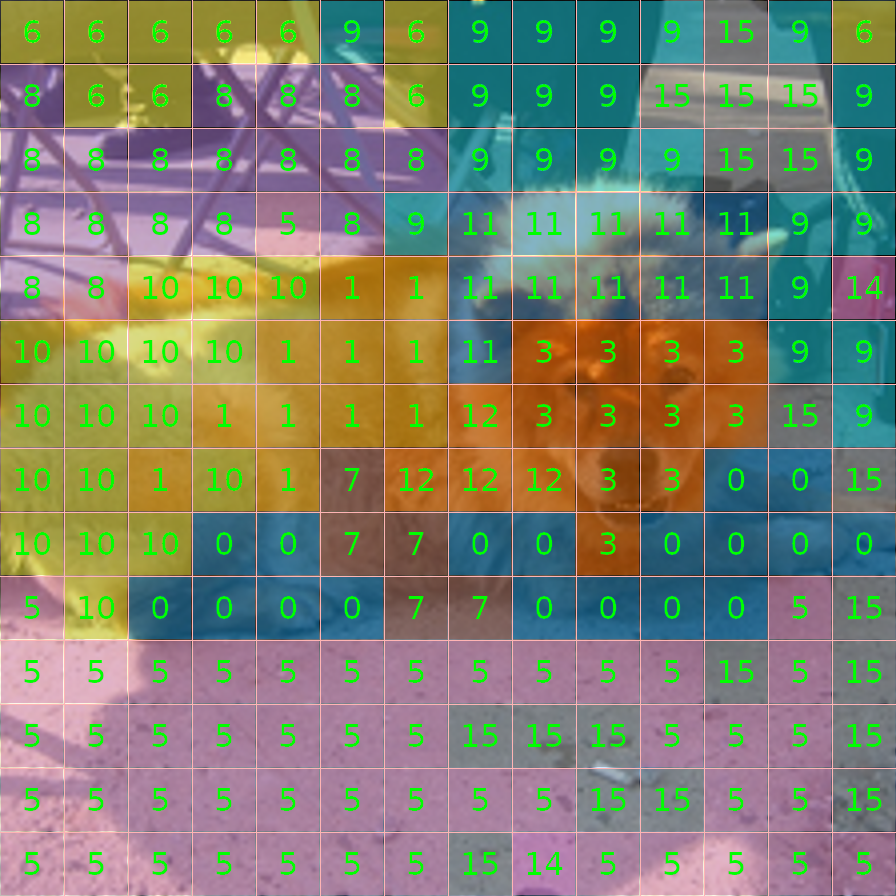} & 
\includegraphics[width=0.15\textwidth]{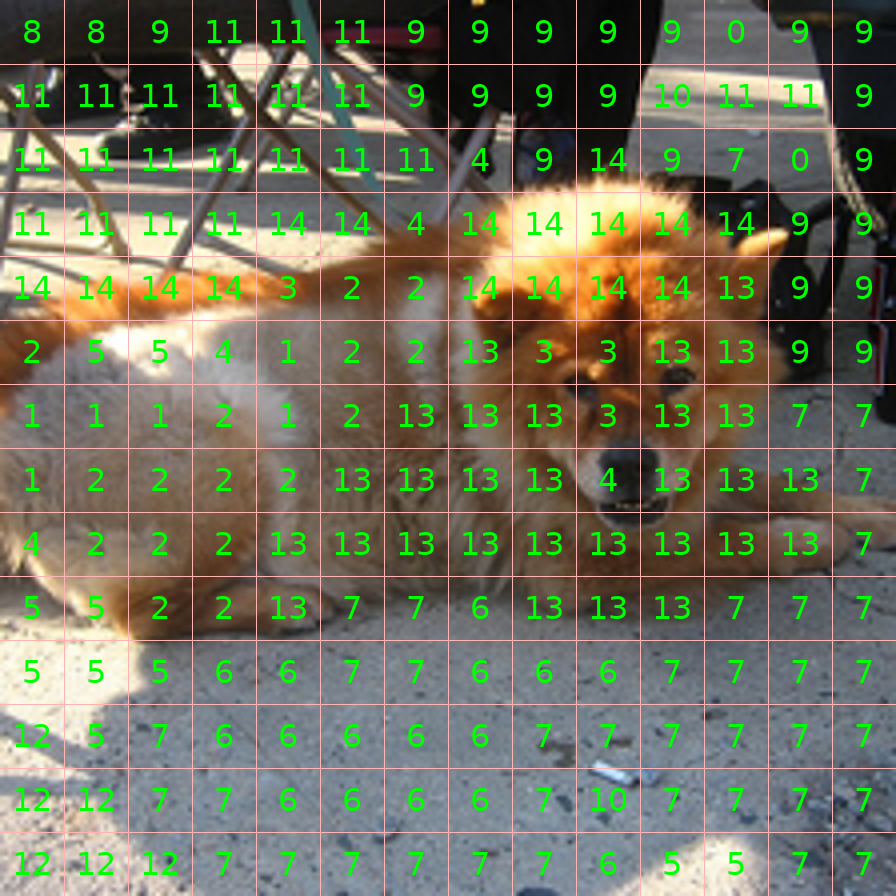} & 
\includegraphics[width=0.15\textwidth]{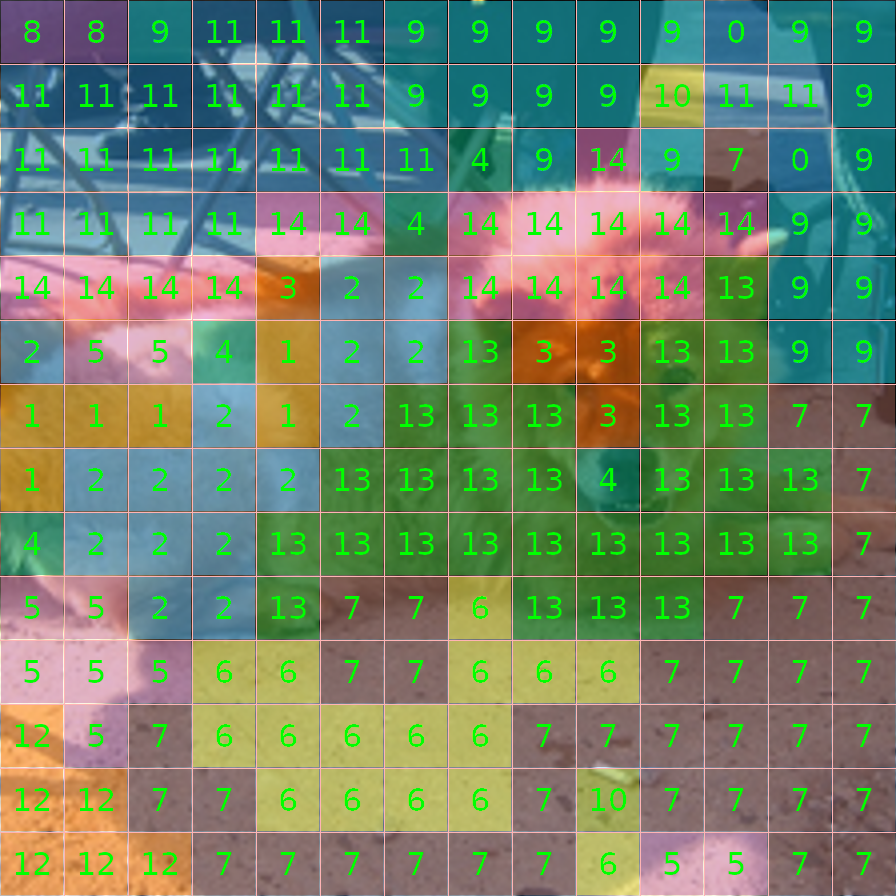} &
\includegraphics[width=0.15\textwidth]{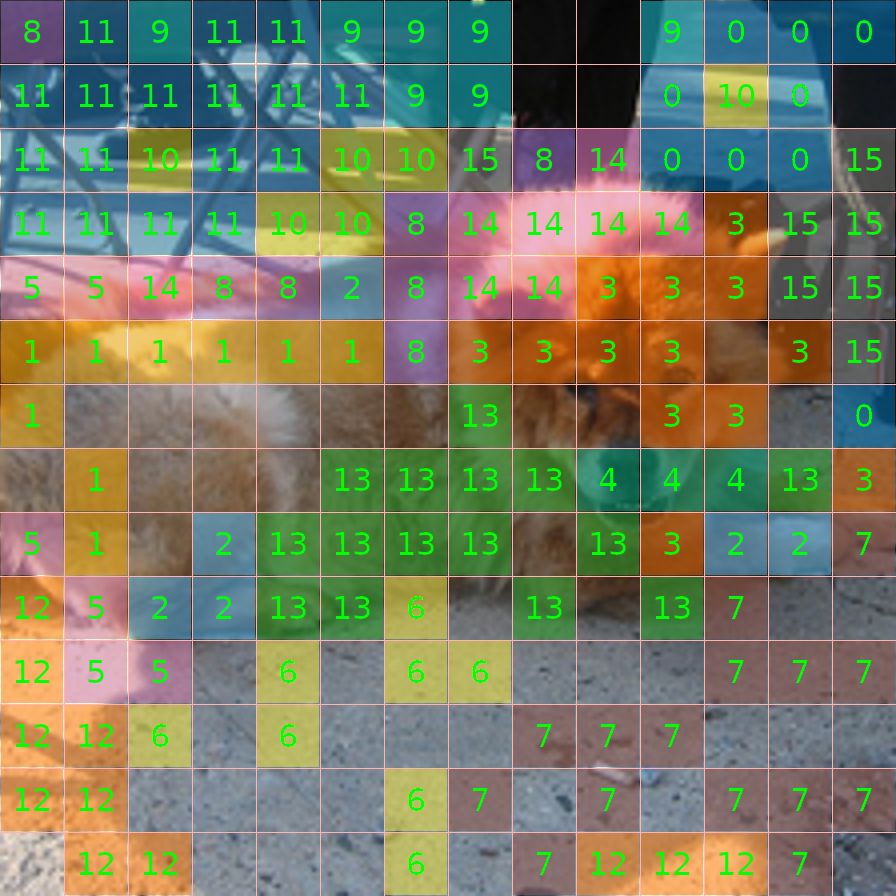} &
\includegraphics[width=0.15\textwidth]{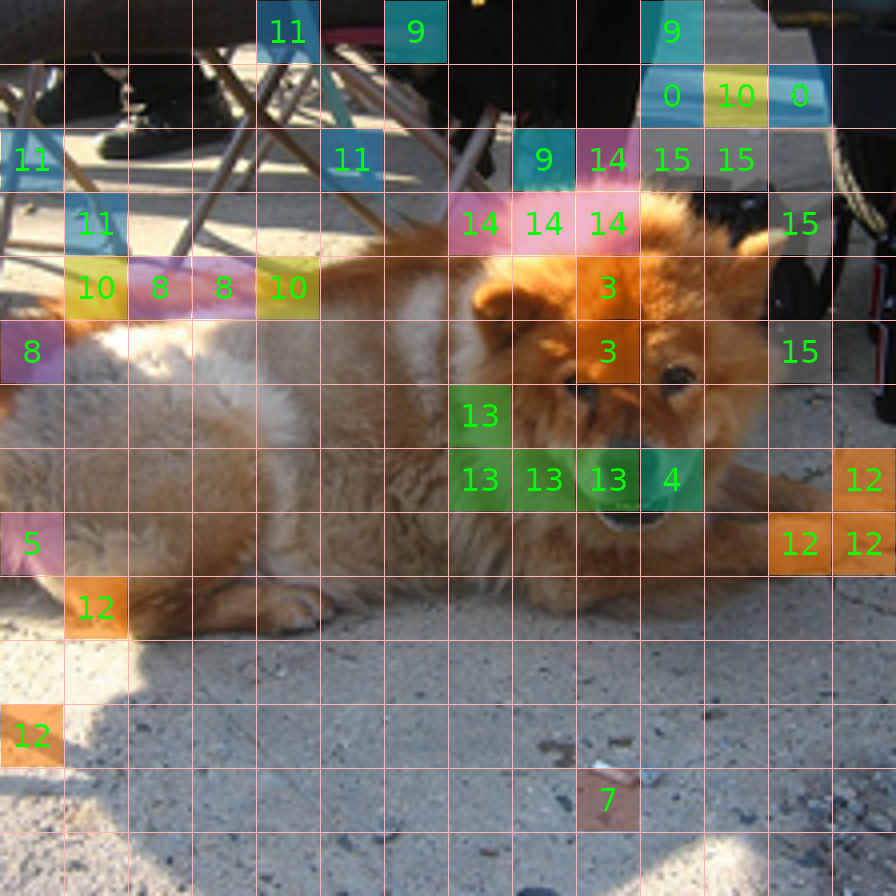}\\
\includegraphics[width=0.15\textwidth]{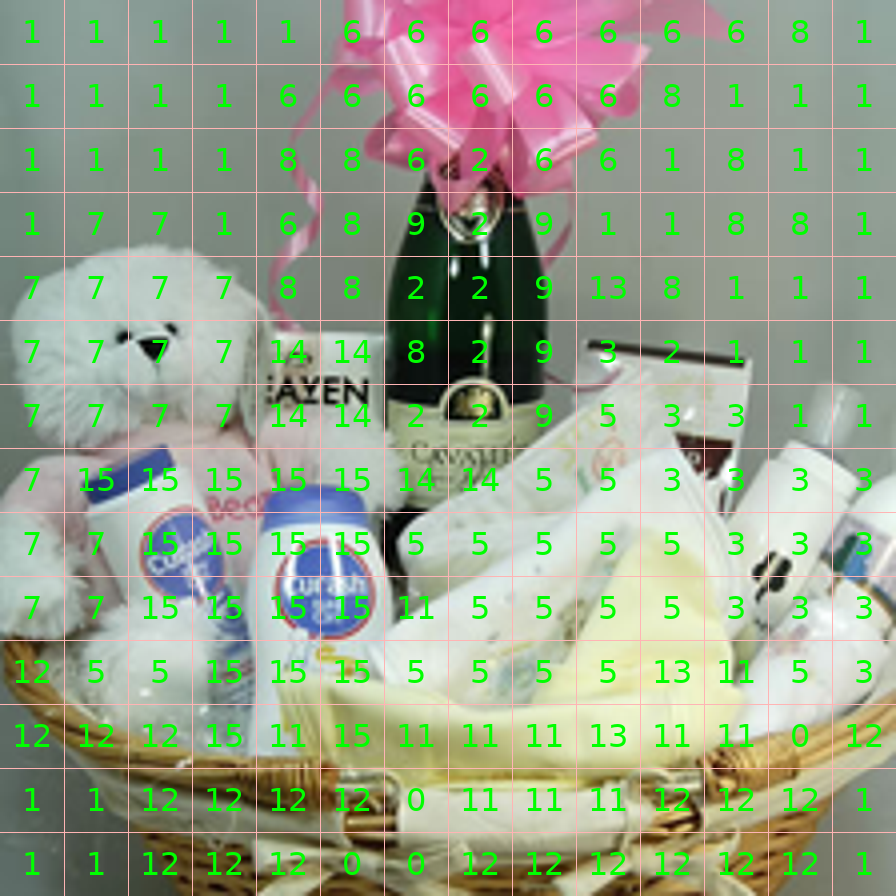} &
\includegraphics[width=0.15\textwidth]{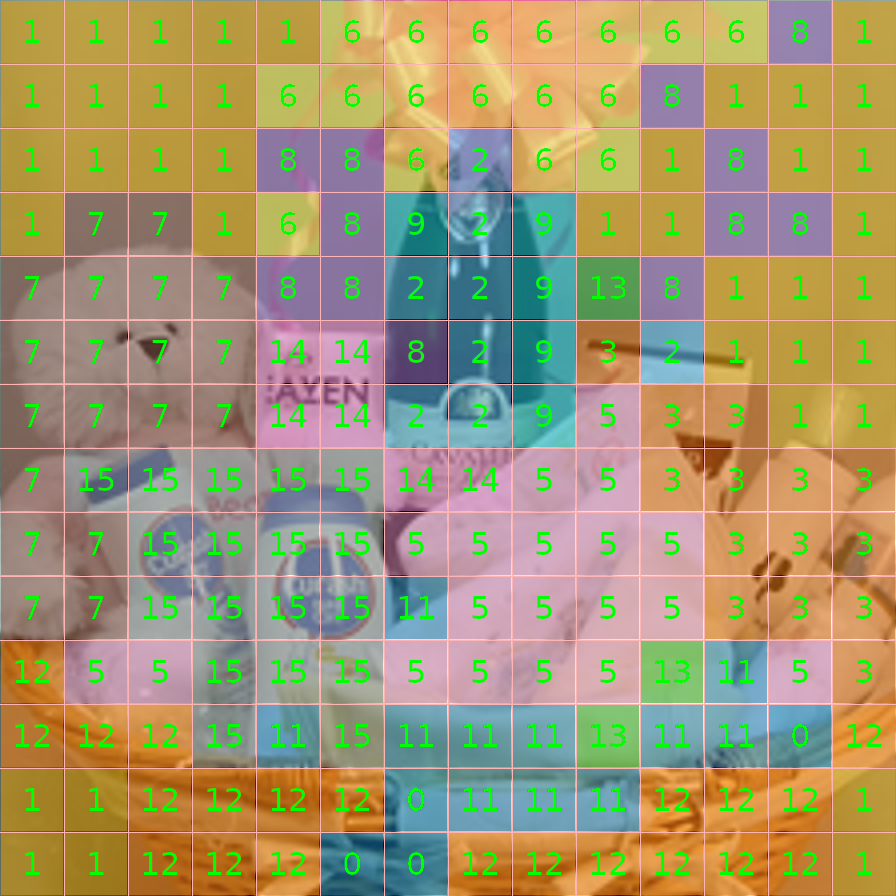} & 
\includegraphics[width=0.15\textwidth]{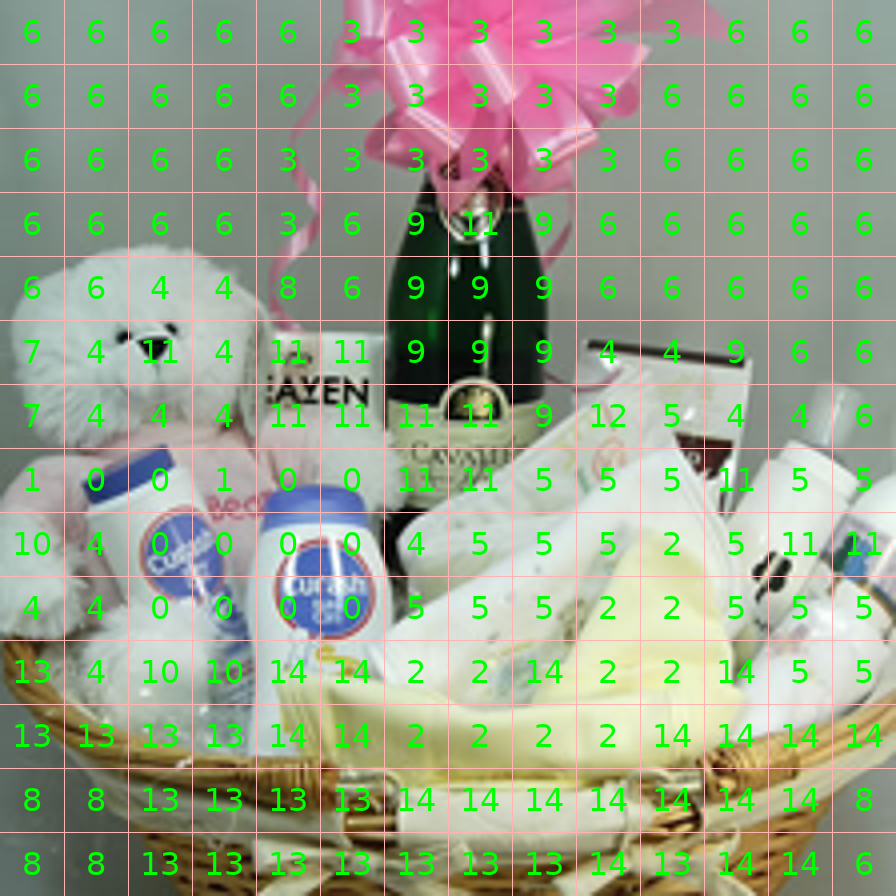} & 
\includegraphics[width=0.15\textwidth]{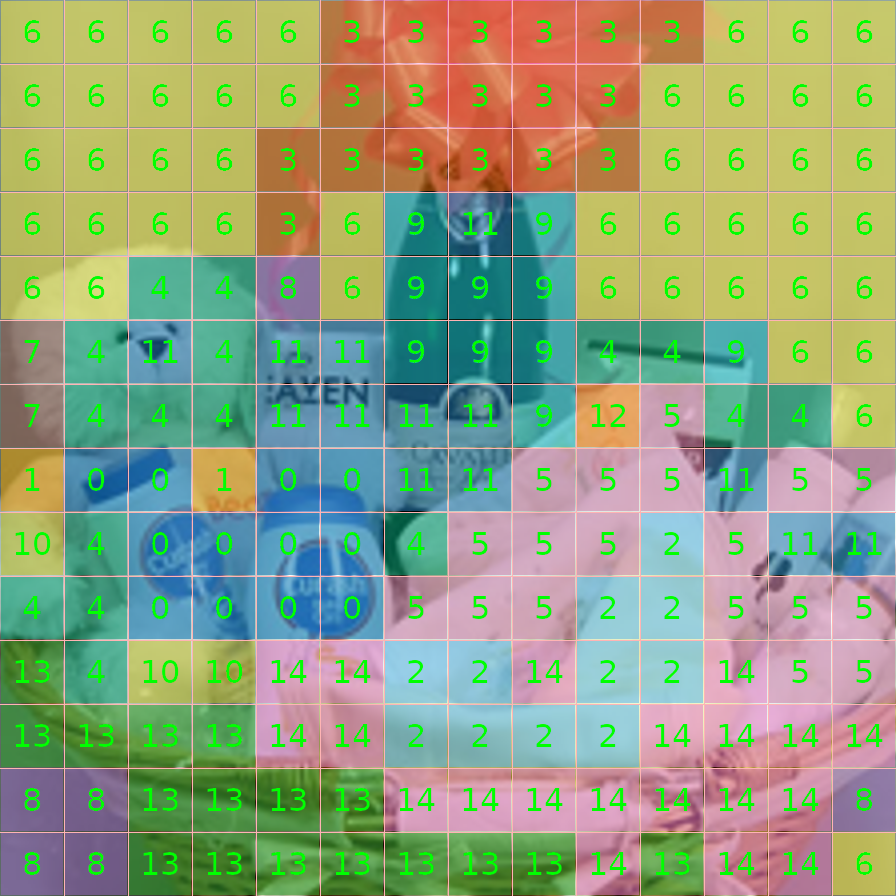} &
\includegraphics[width=0.15\textwidth]{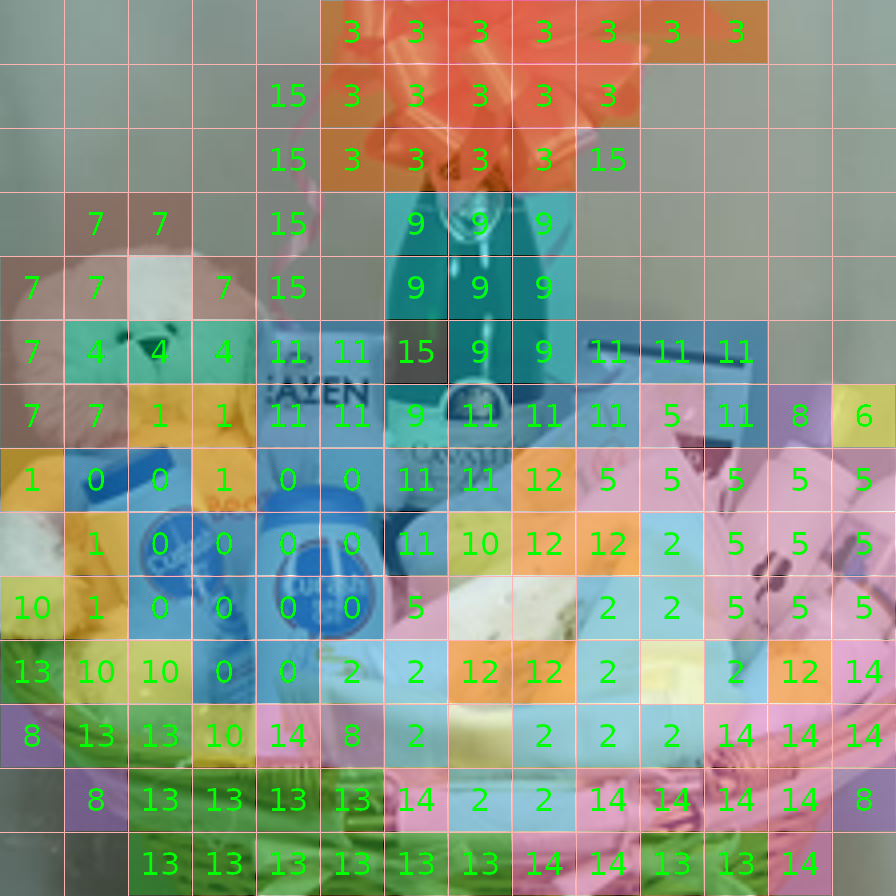} &
\includegraphics[width=0.15\textwidth]{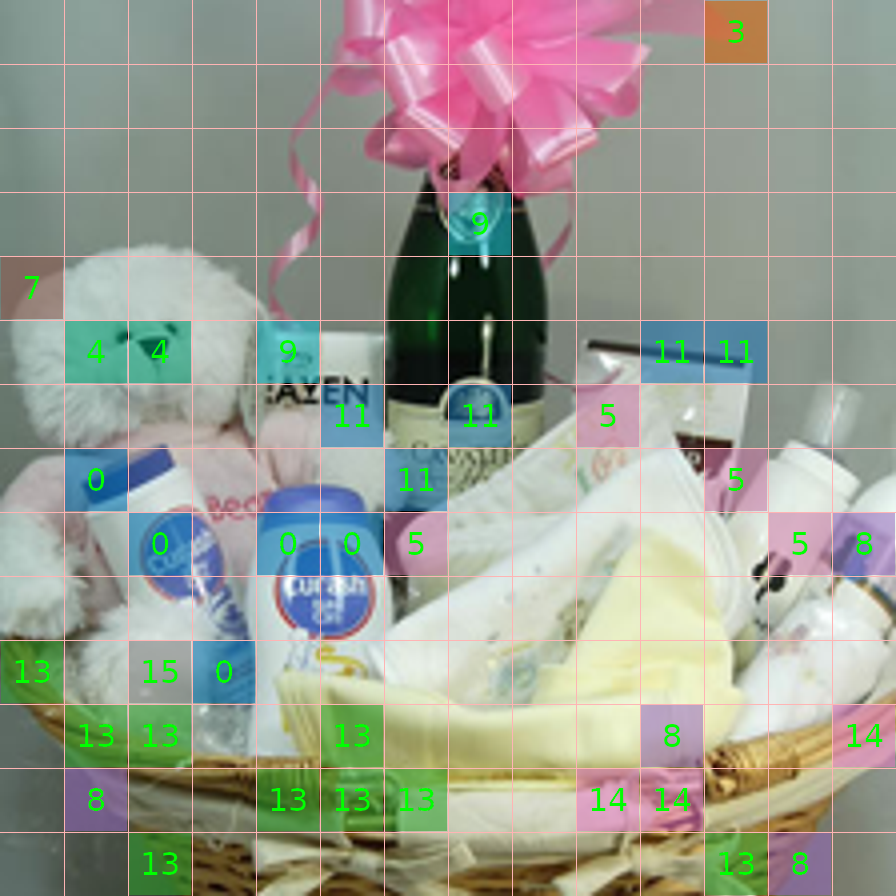}\\
\includegraphics[width=0.15\textwidth]{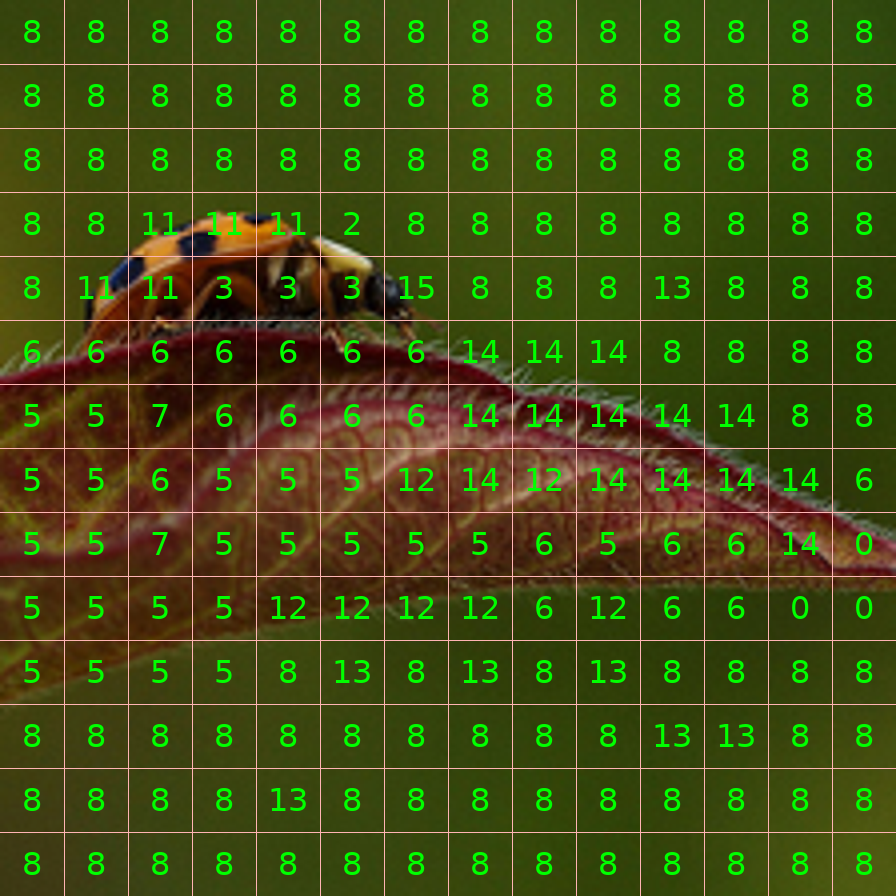} &
\includegraphics[width=0.15\textwidth]{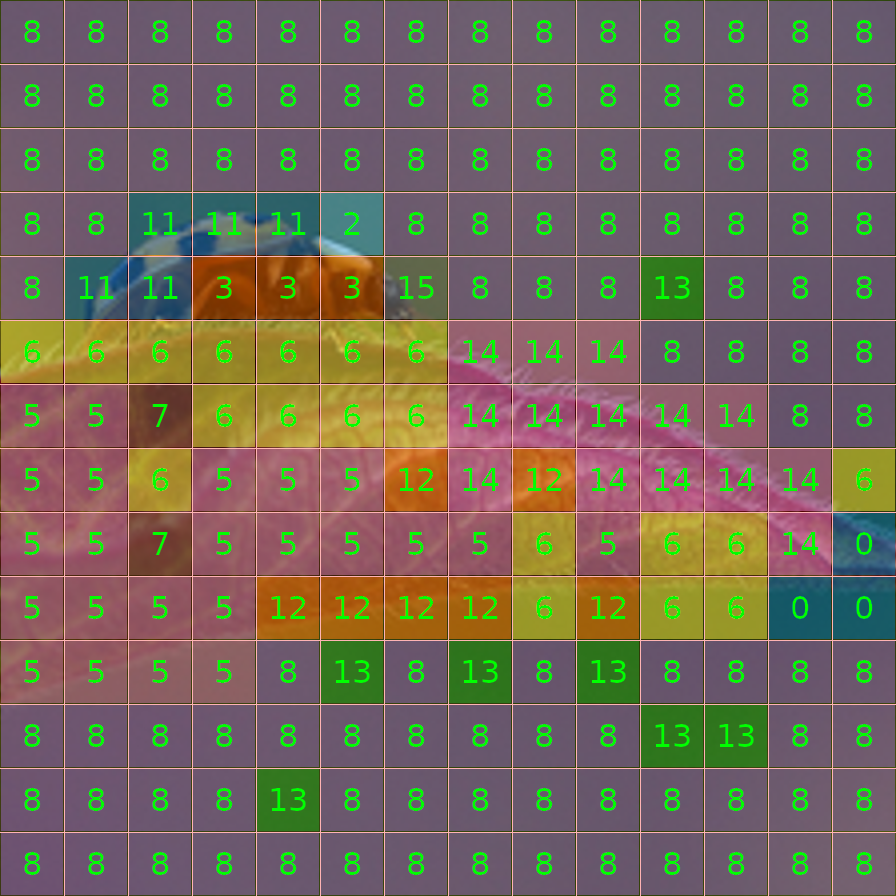} & 
\includegraphics[width=0.15\textwidth]{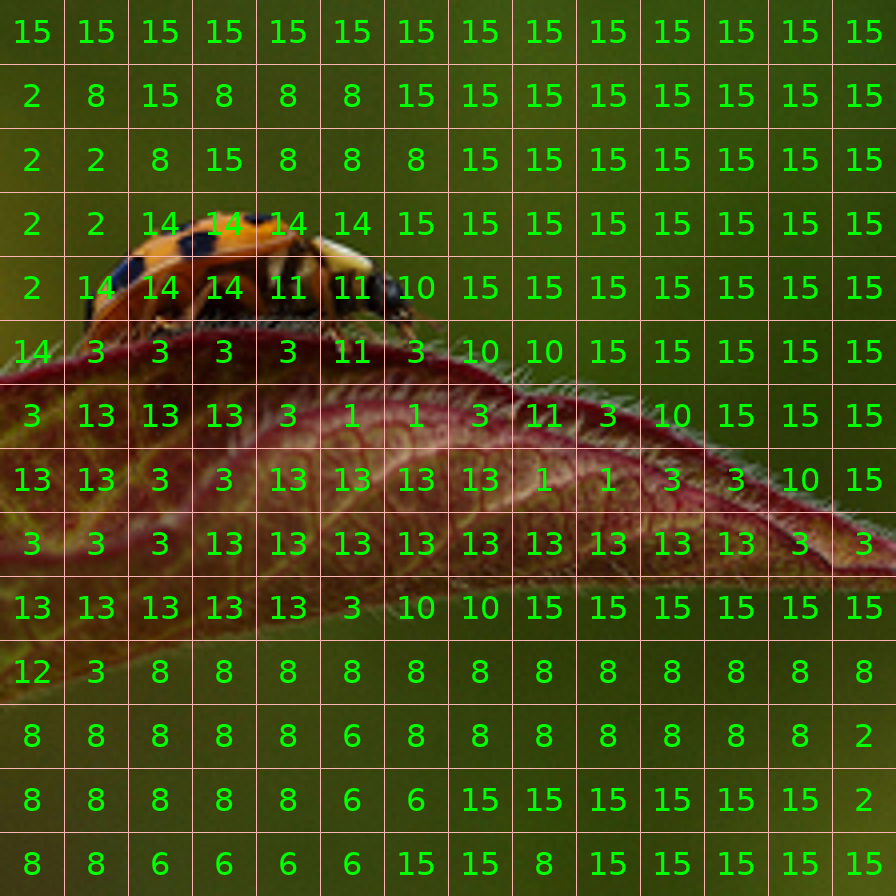} & 
\includegraphics[width=0.15\textwidth]{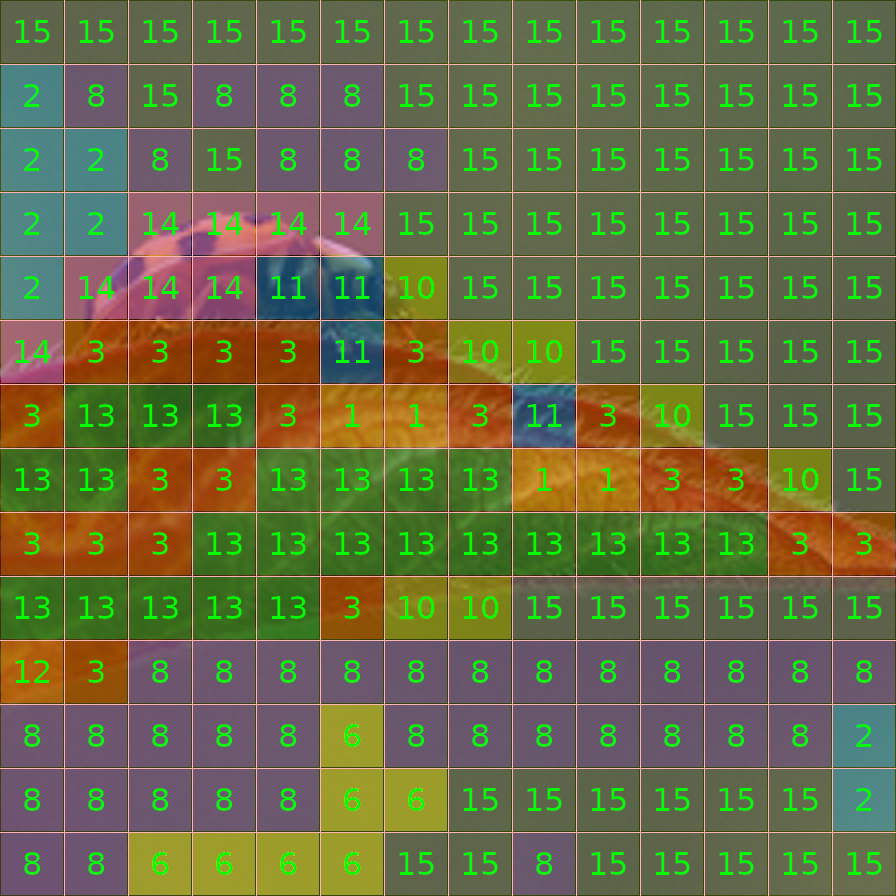} &
\includegraphics[width=0.15\textwidth]{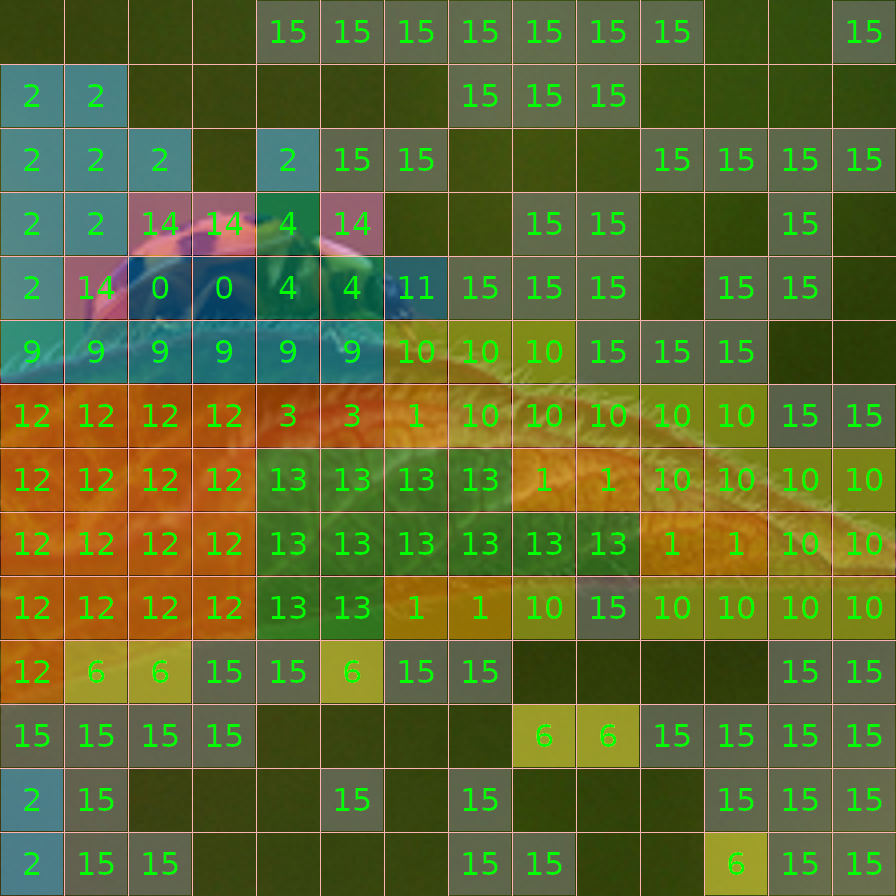} &
\includegraphics[width=0.15\textwidth]{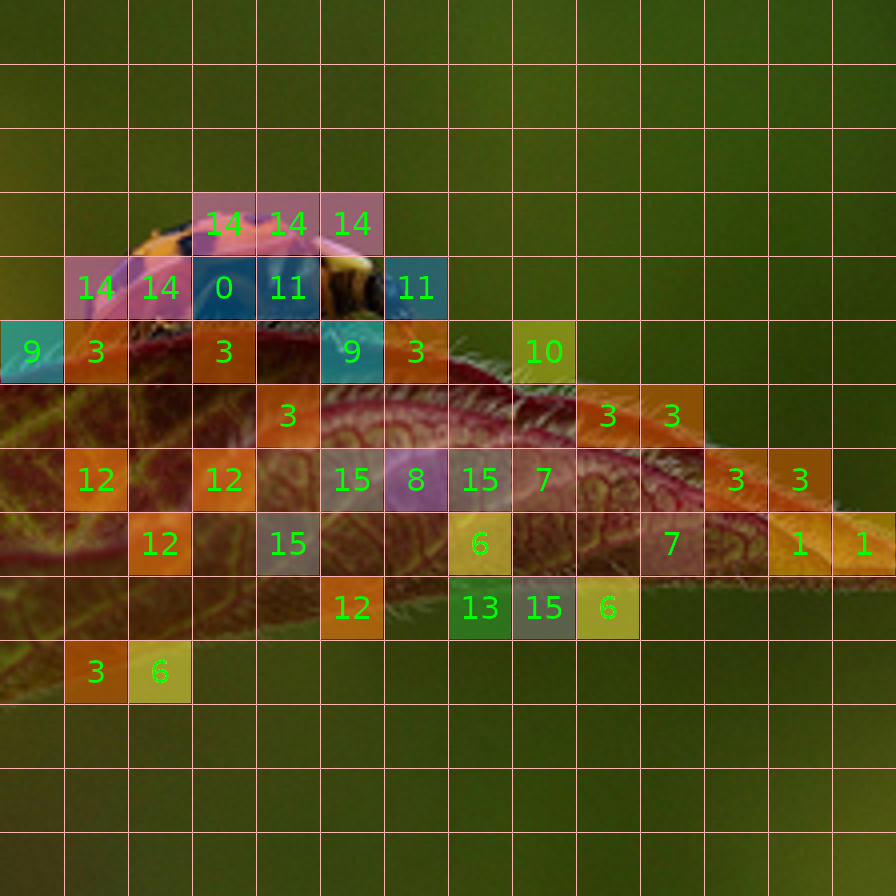}\\
\includegraphics[width=0.15\textwidth]{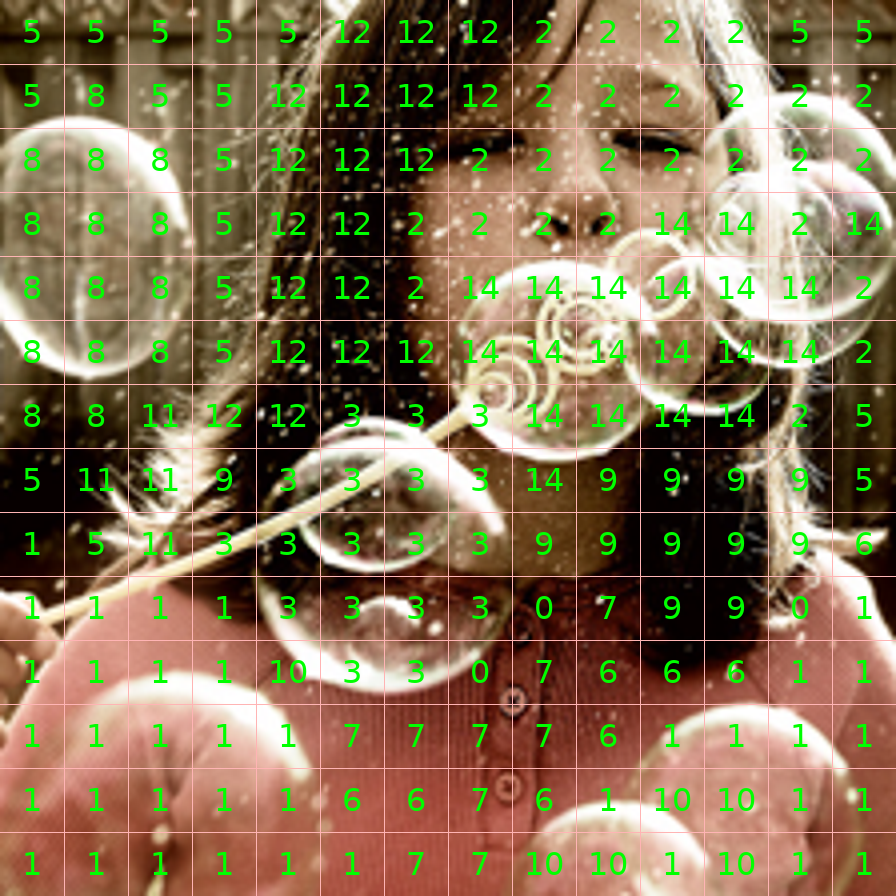} &
\includegraphics[width=0.15\textwidth]{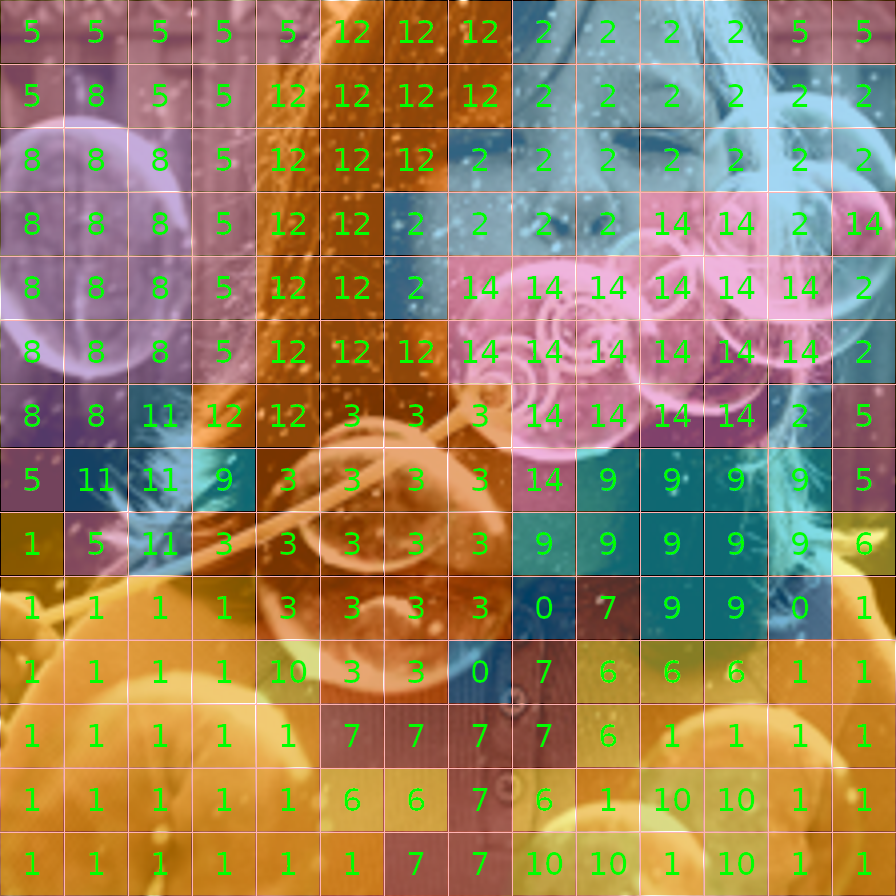} & 
\includegraphics[width=0.15\textwidth]{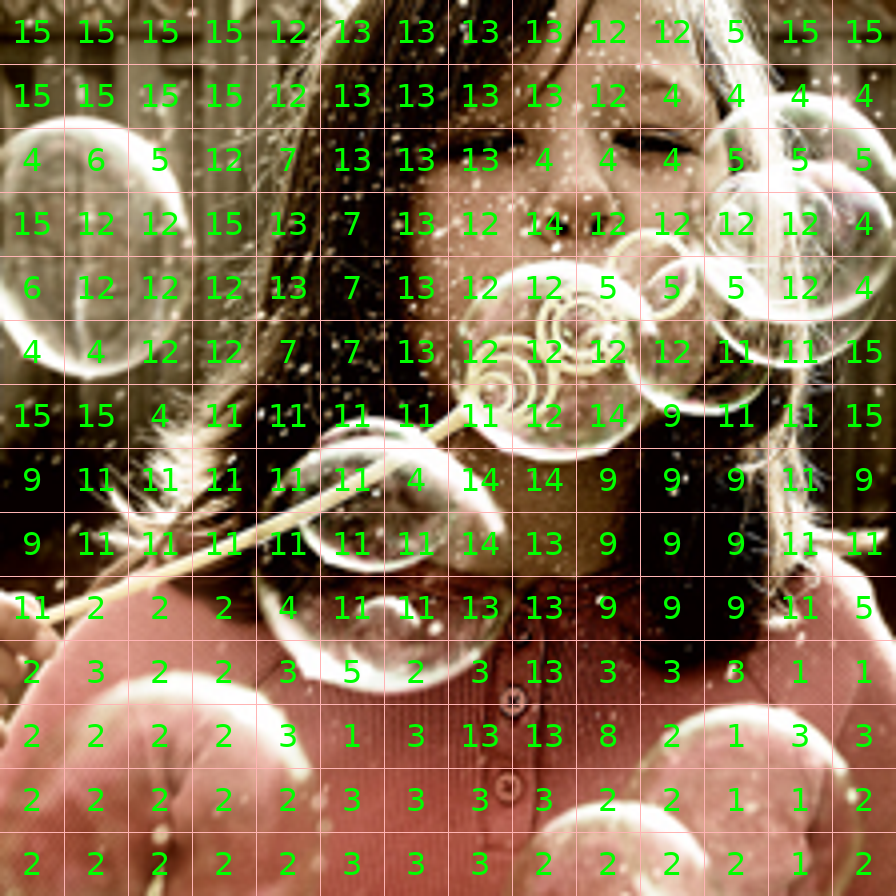} & 
\includegraphics[width=0.15\textwidth]{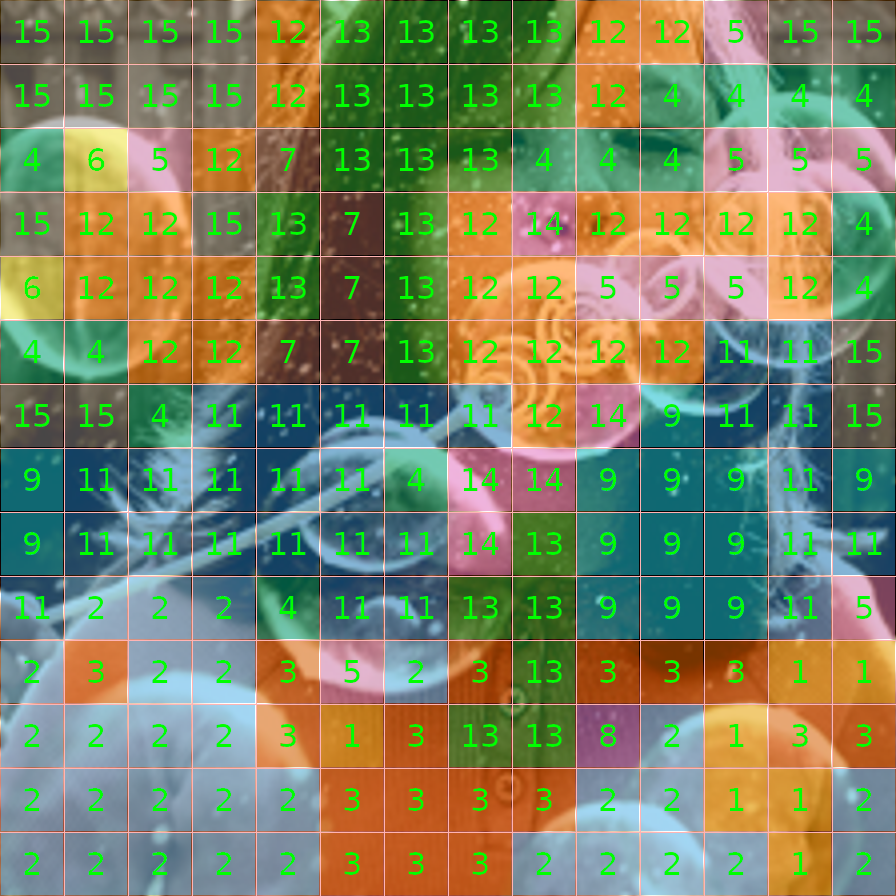} &
\includegraphics[width=0.15\textwidth]{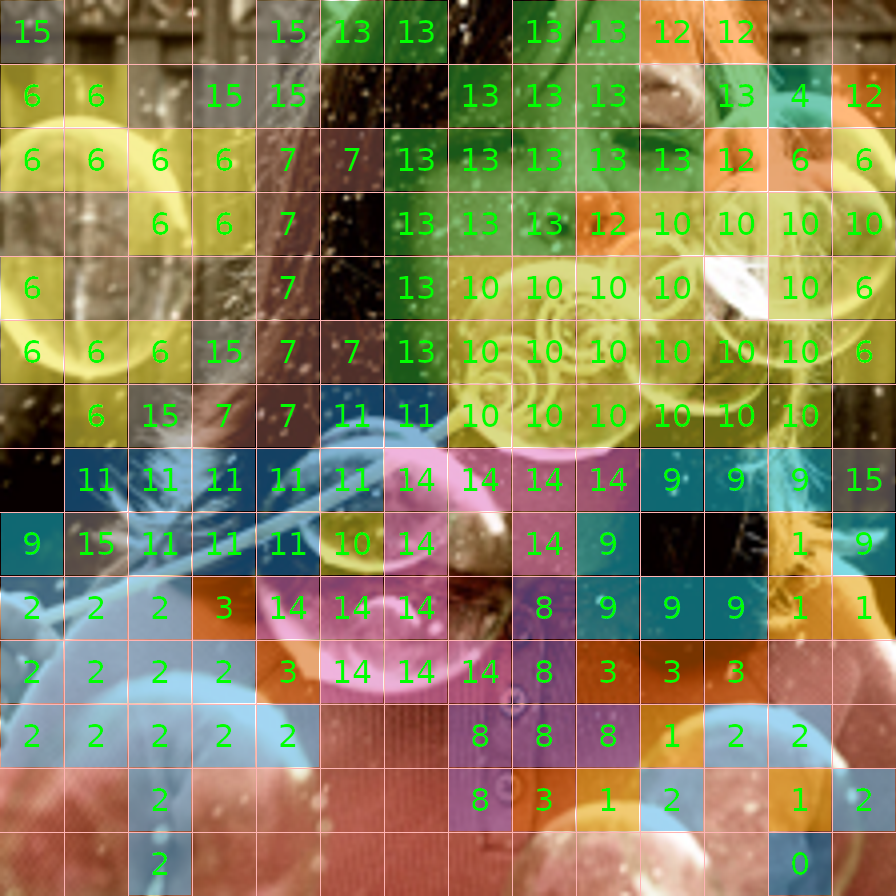} &
\includegraphics[width=0.15\textwidth]{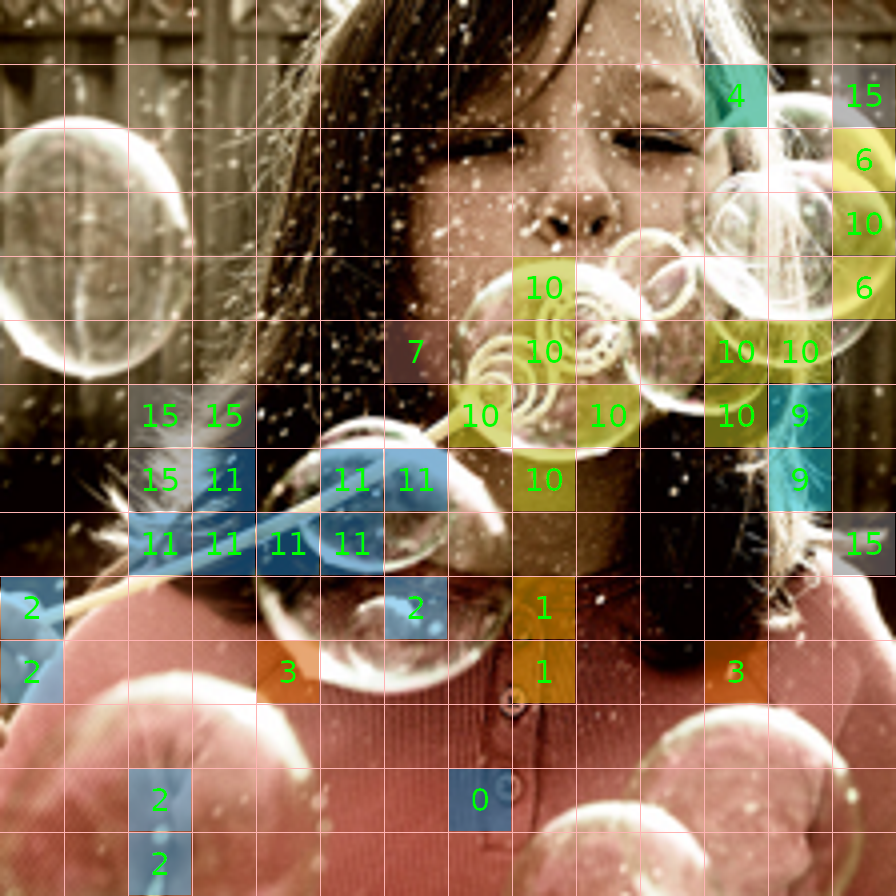}\\
\end{tabular}
\endgroup
\caption{Patch-to-SMB assignments in our single-layer A5 classification model with a keep ratio of $0.0$ and our three-layer A1A3A5 classification model with keep ratios of $0.75$, $0.25$, and $0.0$. From left to right, the figure shows the patch-to-SMB assignments and corresponding color-coded masks at the single compression layer of the A5 model, followed by the patch-to-SMB assignments for the tokens entering the first, second, and third compression layers of the A1A3A5 model. For example, the visualization at the first layer therefore includes all input patch tokens; this layer subsequently drops $25\%$ of them, and the retained tokens enter the second compression layer. Green numbers identify the SMB tokens assigned to reconstruct the corresponding patches, with patches assigned to the same SMB sharing the same color. Patches without a number have been dropped by preceding compression layers. The examples are randomly selected from the ImageNet validation set.}
\label{fig:smb_vis_cls_apdx2}
\end{figure*}

\paragraph{Ablations}\label{cls_ablations}

\begin{table}[t]
\centering
\caption{Ablation studies on model components and architectural design choices.}
\label{tab:cls_ablations}
\small
\setlength{\tabcolsep}{3.5pt}
\renewcommand{\arraystretch}{1.08}
\begin{tabular}{l l c}
\toprule
Model & Ablation & Accuracy \\
\midrule
A1A3A5-SMB16-$0.75,0.25,0.0$ & -- & 79.92 \\
A1A3A5-SMB16-$0.75,0.25,0.0$ & return residual tokens & 79.93 \\
A1A3A5-SMB16-$0.75,0.25,0.0$ & no decoder & 75.95 \\
A1A3A5-SMB16-$0.75,0.25,0.0$ & random token dropping & 79.21 \\
A1A3A5-SMB16-$0.75,0.25,0.0$ & decoder without positional input & 79.91 \\
A1A3A5-SMB16-$0.75,0.25,0.0$ & late SMB injection & 79.70 \\
\midrule
A5-SMB16-$0.0$ & -- & 80.79 \\
A5-SMB16-$0.0$ & late SMB injection & 79.59 \\
\bottomrule
\end{tabular}
\end{table}

We conduct ablation studies to evaluate the contribution of each model component and architectural choice (Table~\ref{tab:cls_ablations}). We use the A1A3A5 configuration with keep ratios of $0.75$, $0.25$, and $0.0$ as the default setting for ablations, which achieves 79.92\% top-1 accuracy on ImageNet-1K. First, replacing the outputs of the compressive layers with residual tokens instead of original tokens does not change accuracy, suggesting that the retained representation is already sufficient for downstream classification. 

Next, removing the decoder entirely while keeping the SMB token stream leads to an approximately 4\% drop in accuracy, indicating that the reconstruction objective is important for learning useful SMB tokens. When the decoder and reconstruction loss are preserved, replacing residual-norm-based token dropping with random dropping results in a modest 0.7\% drop, suggesting that residual norms provide a useful, though not perfect, proxy for token importance. We also examine when the SMB stream is introduced. In our default design, SMB tokens are present from the first layer, which slightly increases compute. As an alternative, we inject learnable SMB tokens only at the first compressive layer. This causes a 0.2\% accuracy drop in the A1A3A5 setting, but a larger 1.2\% drop in the A5 setting, where compression begins later in the network.

Finally, we replace the positional embedding input to the decoder with a constant zero vector, effectively learning a per-SMB prototype. This change has little effect on classification accuracy, even though the reconstruction error increases substantially during training. To quantify this effect, we compute the coefficient of determination ($R^2$) for token predictions in the A1A3A5 model with and without positional embeddings in the decoder. With correct positional embeddings, the $R^2$ values are 0.4827, 0.4814, and 0.5752 for the three compressive layers, respectively. These values decrease to 0.2741, 0.2669, and 0.3168 when the decoder always receives the zero vector as positional input.

\subsection{MLLM}\label{appendix:MLLM}

\paragraph{Implementation Details.}
Following LLaVA-1.5, we use AdamW with a cosine learning-rate schedule. The only modifications are the inserted compressive layers and the additional reconstruction loss, which is applied during both pretraining and fine-tuning with different weights to balance compression and task adaptation. We keep the rest of the training recipe unchanged and train the two modules jointly for 2 epochs. For evaluation, we use greedy decoding with temperature set to 0 to ensure deterministic outputs.

\paragraph{Datasets.}
We use the 558K image-text alignment dataset and the 665K multimodal instruction dataset released with the backbone~\cite{liu2024improved} for pretraining and instruction tuning, respectively. We evaluate our model on seven widely adopted benchmarks, including MMBench~\cite{liu2024mmbench}, MME~\cite{fu2023mme}, POPE~\cite{li2023evaluating}, SQA~\cite{lu2022learn}, VQAText~\cite{singh2019towards}, and VQA~\cite{goyal2017making}.

\paragraph{Multi-model Understanding}~\label{sec:result-MLLM-normal}
We evaluate the proposed compression framework on multimodal reasoning and understanding benchmarks using LLaVA-1.5-13B~\cite{liu2024improved} under retained token budgets of 128, 64, and 32. We include ToMe~\cite{bolya2023tome} as a related baseline, although its performance is relatively limited in the MLLM setting, since it represents an early attempt to bridge ViT compression and multimodal models. For MMBench and MMBench(CN)~\cite{liu2024mmbench}, we report dev-set results because test labels are not publicly available and the online evaluation server is no longer accessible.

As shown in Table~\ref{tab:llava13b_results}, SMB remains competitive across token budgets while substantially reducing the visual prefix length. It is particularly effective on reasoning-heavy benchmarks such as MMB and MMB-CN, and it continues to preserve most of the baseline capability with 64 or even 32 retained tokens, indicating that the learned compression is robust and scalable. On POPE, the relatively small gap on the adversarial split suggests that token compression has limited additional impact on robustness to semantically plausible absent objects, while the lower recall indicates that the compressed model remains conservative in predicting object presence.

\begin{table*}[t]
\centering
\caption{Comparison of token retention methods on LLaVA-1.5-13B across varying token budgets. Raw scores are reported in the first row and average percentages to the baseline in the final column.  Results for SparseVLM and ToMe are sourced from~\cite{zhang2024sparsevlm}. Comparisons are best interpreted via relative percentages. The baseline scores 76.4 and 72.4 on the dev splits of MMB and MMB$^{\mathrm{CN}}$, respectively}
\label{tab:llava13b_results}
\small
\setlength{\tabcolsep}{2.2pt}
\renewcommand{\arraystretch}{1.16}
\begin{tabular*}{\textwidth}{@{\extracolsep{\fill}}%
L{3.35cm}@{\hspace{-4pt}}
C{0.78cm}
C{0.80cm}
C{0.84cm}
C{0.78cm}
C{0.66cm}
C{0.66cm}
C{0.70cm}
C{0.66cm}}
\toprule
\textbf{Method}
& \textbf{VQA$^{v2}$}
& \textbf{SQA$^{\mathrm{IMG}}$}
& \textbf{TextVQA}
& \textbf{POPE}
& \textbf{MME}
& \textbf{MMB}
& \textbf{MMB$^{\mathrm{CN}}$}
& \textbf{Avg.} \\
\midrule
\multicolumn{9}{c}{{\itshape Vanilla 576 Tokens}} \\
LLaVA-1.5-13B
& 80.0
& 71.6
& 61.3
& 86.0
& 1531
& 67.7
& 63.5
& 100\% \\
\midrule
\multicolumn{9}{c}{{\itshape Retain 128 Tokens}} \\
ToMe(ICLR'23)~\citenumonly{bolya2023tome}
& -
& 59.6
& 49.1
& 62.8
& 1343
& 53.3
& -
& 79.6\% \\
FastV\citenumonly{chen2024image}(ECCV'24)
& 75.3
& 74.2
& 58.6
& 75.5
& 1460
& 66.1
& 62.3
& 96.0\% \\
PDrop\citenumonly{xing2024pyramiddrop}(CVPR'25)
& 78.2
& 73.3
& 60.2
& 83.6
& 1489
& 67.5
& 62.8
& 98.4\% \\
SparseVLM\citenumonly{zhang2024sparsevlm}(ICML'25)
& -
& 68.6
& 56.7
& 85.0
& 1488
& 64.5
& -
& 97.7\% \\
PruMerge+\citenumonly{shang2025llava}(ICCV'25)
& 76.2
& 73.3
& 56.1
& 82.7
& 1446
& 66.3
& 61.2
& 95.8\% \\
VisionZip\citenumonly{yang2025visionzip}(CVPR'25)
& 76.8
& 73.8
& 58.9
& 82.7
& 1450
& 67.4
& 62.5
& 96.7\% \\
ReConV(Ours)
& 75.0
& 70.9
& 55.3
& 82.2
& 1457
& 74.0
& 71.5
& 95.8\% \\
\midrule
\multicolumn{9}{c}{{\itshape Retain 64 Tokens}} \\
ToMe\citenumonly{bolya2023tome}(ICLR'23)
& -
& 50.0
& 45.3
& 52.5
& 1138
& 43.7
& -
& 67.9\% \\
FastV\citenumonly{chen2024image}(ECCV'24)
& 65.3
& 73.1
& 53.4
& 56.9
& 1246
& 59.2
& 55.1
& 85.8\% \\
PDrop\citenumonly{xing2024pyramiddrop}(CVPR'25)
& 70.8
& 73.1
& 55.3
& 66.1
& 1247
& 63.1
& 56.6
& 88.7\% \\
SparseVLM\citenumonly{zhang2024sparsevlm}(ICML'25)
& -
& 69.8
& 53.4
& 77.5
& 1373
& 60.1
& -
& 92.1\% \\
% SparseVLM (ICML'25)
% & 73.2
% & 73.0
% & 57.1
% & 77.9
% & 1374
% & 65.2
% & 60.3
% & 93.5\% \\
PruMerge+\citenumonly{shang2025llava}(ICCV'25)
& 72.6
& 73.5
& 54.4
& 75.7
& 1338
& 65.0
& 59.3
& 92.3\% \\
VisionZip\citenumonly{yang2025visionzip}(CVPR'25)
& 73.7
& 74.2
& 57.4
& 75.7
& 1380
& 64.9
& 61.3
& 93.9\% 
\\
ReConV(Ours)
& 72.5
& 68.5
& 52.9
& 82.6
& 1404
& 73.1
& 68.5
& 92.6\% \\
\midrule
\multicolumn{9}{c}{{\itshape Retain 32 Tokens}} \\
ReConV(Ours)
& 70.9
& 70.6
& 52.8
& 82.1
& 1352
& 72.6
& 69.0
& 92.5\% \\
\bottomrule
\end{tabular*}
\end{table*}

\paragraph{Efficiency.}
We measure prefill latency on POPE on a single A800 GPU, as reported in Table~\ref{tab:pope_latency_efficiency}. In the 64-token setting, our method achieves a prefill latency of 37.84 ms per sample, corresponding to a $2.97\times$ speedup over the backbone, while retaining 96\% of the backbone performance on POPE. Because compression shortens the visual prefix during both prefill and training, it also yields a comparable reduction in training and fine-tuning time.

\subsection{Semantic Partition Map in MLLM}\label{sec:apdx-mllm-smb-vis}
\begin{figure}[t]
    \centering
    \includegraphics[width=\columnwidth]{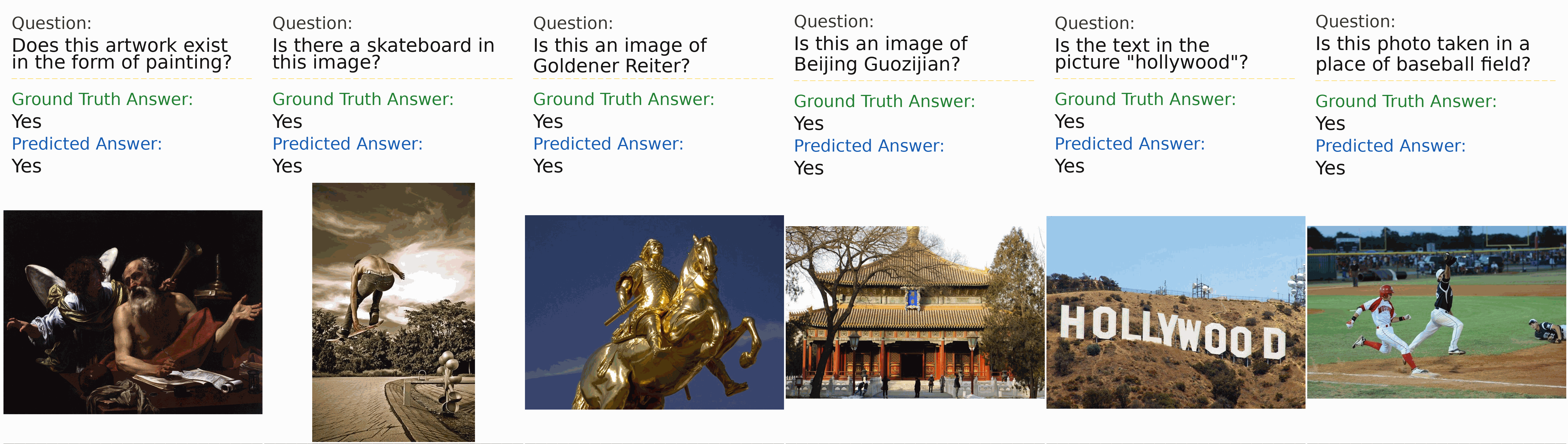}\\[-2pt]
    \includegraphics[width=\columnwidth]{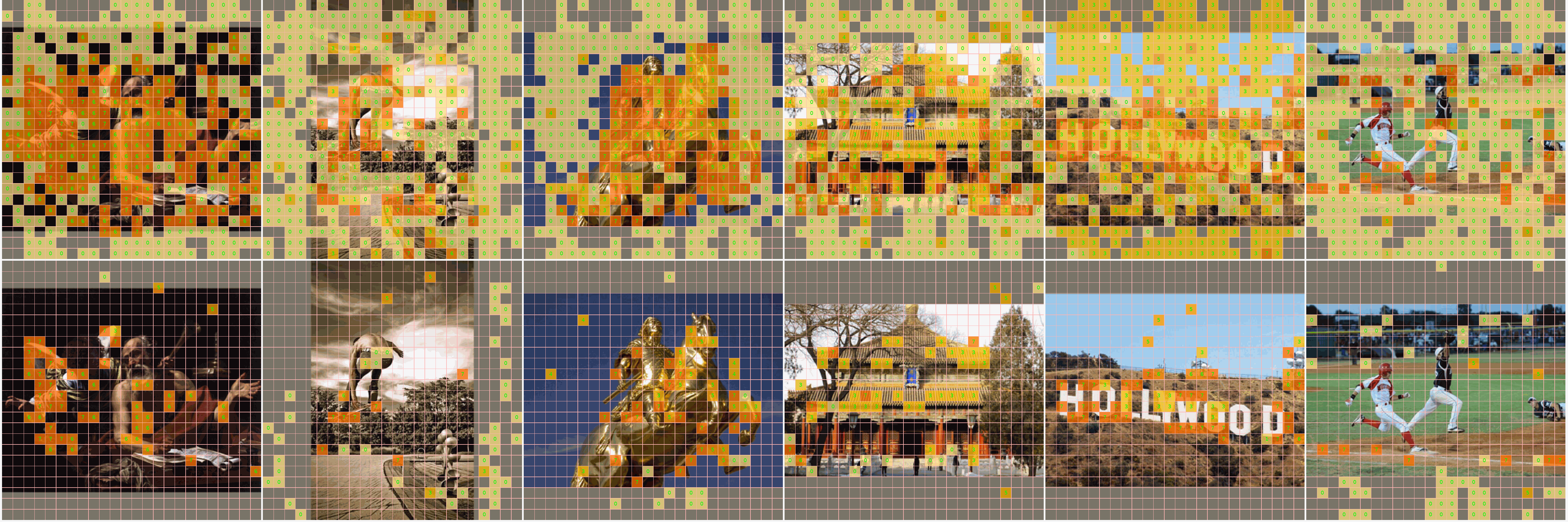}
    \caption{Samples are randomly selected from the MME evaluation set, covering five subtasks from left to right: artwork, existence, landmark, OCR, and scene. The figure visualizes the 64-token setting of our model, which contains 56 kept tokens and 8 SMBs. From top to bottom, the panels show: the question-answer pair, the input image, SMB allocation at the first compression layer, SMB allocation at the second compression layer after the first token-dropping step, and the final SMB allocation after the second token-dropping step. The color from light to dark yellow indicates SMB indices from 0 to 8. The letters on each patch indicate the assigned SMB index, while patches without letters have already been dropped.}
    \label{fig:MMEsmbPartition}
\end{figure}
The visualization of SMB partitioning further shows that after two compression layers, the retained visual tokens align well with the keywords in the question prompt, indicating that the kept tokens preserve the fine-grained visual details relevant to downstream reasoning and remain semantically grounded in the task-specific query. Meanwhile, SMB provides complementary global semantic context. Together, keyword-level detail preservation and global semantic coverage help explain the effectiveness of our method under visual self-driven signals, without relying on attention saliency.

\begin{table}[t]
\centering
\caption{Efficiency analysis on POPE. 
We report prefill latency per sample, measured on an A800 GPU 
(40GB). Speedup is computed relative to the Baseline.}
\label{tab:pope_latency_efficiency}
\small
\setlength{\tabcolsep}{4pt}
\renewcommand{\arraystretch}{1.08}
\begin{tabular}{l c c c c c c}
\toprule
\multirow{2}{*}{Method} & \multirow{2}{*}{Tokens} & \multicolumn{3}{c}{POPE Accuracy} & \multicolumn{1}{c}{Prefill Latency} & \multicolumn{1}{c}{Speedup} \\
\cmidrule(lr){3-5}\cmidrule(lr){6-6}\cmidrule(lr){7-7}
& & rand & pop & adv & /Sample(ms)$\downarrow$ & (Prefill)$\uparrow$ \\
\midrule
Baseline & 576 & 87.1 & 86.2 & 84.5 & 112.25 & 1.00$\times$ \\
ReConV(Ours) & 127 & 83.3 & 82.8 & 80.4 & 49.63 & 2.26$\times$ \\
ReConV(Ours)  & 64  & 84.5 & 82.9 & 80.3 & 37.84 & 2.97$\times$ \\
ReConV(Ours)  & 32  & 84.8 & 82.5 & 78.9 & 36.55 & 3.07$\times$ \\
\bottomrule
\end{tabular}
\end{table}

\end{document}